\documentclass[journal]{IEEEtran}

\ifCLASSINFOpdf
\else

\fi

\usepackage{amsmath,amsfonts}
\usepackage{algorithmic}
\usepackage{algorithm}
\usepackage{array}
\usepackage[caption=false,font=normalsize,labelfont=sf,textfont=sf]{subfig}
\usepackage{textcomp}
\usepackage{stfloats}
\usepackage{url}
\usepackage{verbatim}
\usepackage{enumitem}
\usepackage{graphicx}
\usepackage{cite}
\usepackage{makecell,rotating,multirow,diagbox}
\usepackage{subfloat}

\usepackage{booktabs}
\usepackage{threeparttable}
\usepackage[table]{xcolor}

\newcommand{\qed}{\hfill $\Box$ }
\newtheorem{theorem}{Theorem}[section]

\newtheorem{remark}{Remark}[section]

\begin{document}
%
\title{Breaking Spatial Boundaries: Spectral-Domain Registration Guided Hyperspectral and Multispectral Blind Fusion}
%
%
%

\author{Kunjing Yang,~
        Libin Zheng,~
        Minru Bai,~
        Ting Lu,~\IEEEmembership{Member, ~IEEE,~}
        Leyuan Fang,~\IEEEmembership{Senior Member, ~IEEE}
\thanks{Kunjing Yang and Libin Zheng are with the School of Mathematics, Hunan University, Changsha, Hunan 410082, P. R. China(e-mail: kunjing-yang@hnu.edu.cn; \quad fifholz301@hnu.edu.cn).}
\thanks{Minru Bai is the Corresponding author with School of Mathematics, Hunan University, Changsha, Hunan 410082, P. R. China (e-mail: minru-bai@hnu.edu.cn).}
\thanks{Ting Lu and Leyuan Fang are with the College of Electrical and
			Information Engineering, Hunan University, Changsha 410082, P. R  China (e-mail: tingluhnu@gmail.com; \quad fangleyuan@gmail.com).}}

\maketitle

\begin{abstract}
The blind fusion of unregistered hyperspectral images (HSIs) and multispectral images (MSIs) has attracted growing attention  recently. To address the registration challenge, most existing methods employ spatial transformations on the HSI to achieve alignment with the MSI. However, due to the substantial differences in spatial resolution of the images, the performance of these methods is often unsatisfactory. Moreover, the registration process tends to be time-consuming when dealing with large-sized images in remote sensing.  To address these issues, we propose tackling the registration problem from the spectral domain. Initially, a lightweight Spectral Prior Learning (SPL) network is  developed to  extract spectral features from the HSI and enhance the spectral resolution of the MSI. Following this, the obtained image undergoes spatial downsampling to produce the  registered HSI. In this process,  subspace representation and  cyclic training strategy are employed to improve spectral accuracy of the registered HSI obtained. Next, we propose a blind sparse fusion (BSF) method, which utilizes group sparsity regularization to equivalently promote the low-rankness of the image. This approach not only circumvents the need for rank estimation, but also reduces computational complexity. Then, we employ the Proximal Alternating Optimization (PAO) algorithm to solve the BSF model, and present its  convergence analysis. Finally, extensive numerical experiments on simulated and real datasets are conducted to verify the effectiveness of our method in registration and fusion. We also demonstrate its efficacy in enhancing classification performance.
\end{abstract}


%
\IEEEpeerreviewmaketitle

\section{Introduction}

\IEEEPARstart{H}{yperspectral} images  can incorporate tens to hundreds of contiguous spectral bands,  granting them the capability to encapsulate a broader swath of the electromagnetic spectrum and providing more nuanced spectral details \cite{L2002}. Given that different materials exhibit unique reflection signatures, the extensive spectral coverage of HSIs enables precise material identification, which is  advantageous in various applications, such as disease monitoring \cite{Lv2021}, anomaly detection \cite{He2023}, and land cover classification \cite{Xie2022}.

	However, there are fundamental physical limitations that inevitably impose a trade-off  between the spatial and spectral resolution in HSIs \cite{highcost}. As a result, HSIs typically exhibit low spatial resolution. Moreover, factors such as sensor noise further degrade the quality of the images, limiting their broader applications \cite{Milad}. 	In comparison, MSIs exhibit higher spatial resolution but lower spectral resolution. Currently, a mainstream approach to enhancing the spatial resolution of HSIs involves fusing them with high spatial resolution MSIs, which can facilitate the extraction of more intricate and detailed information from the scene \cite{L2015}.
	
	In real-world applications, to achieve high-quality HSI-MSI fusion performance,  the HSI and MSI need to be pre-registered. In addition,  the spatial and spectral degradation operators, i.e., the point spread function (PSF) and spectral response function (SRF), need to be estimated in advance. To address registration challenges, most existing methods determine geometric spatial transformations by matching spatial feature points between HSIs and MSIs. Then, the transformations are applied to the HSI to achieve  alignment with the MSI \cite{Ren2023, Zheng2022}. However, this procedure, which we refer to as spatial domain registration, has the following limitations: 1) Due to the significant differences in spatial resolution between HSI and MSI, fewer common feature points can be extracted and matched, which often leads to unsatisfactory registration performance in these approaches \cite{Qu2022}. 2) The registration process can be generally time-consuming when handling large-sized images in  remote sensing \cite{Zhou2020}.  3) The spatial registration process may potentially introduce distortions to the spectral domain of the HSIs.
	As a result of these factors, the spatial domain registration methods often fail to efficiently and accurately register the HSIs and MSIs, thereby affecting the subsequent estimation of degradation operators and the fusion process.

\begin{figure}[t]
		\vspace{-2pt}
		\centering
		\!\!\!\includegraphics[width=9.0cm]{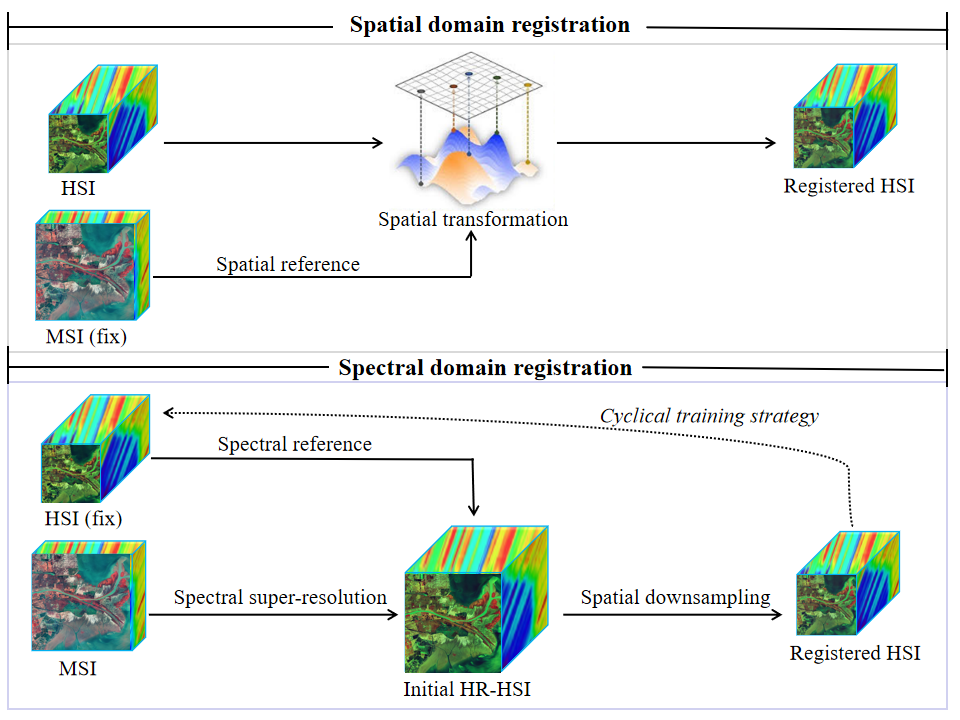}
		
		\vspace{-4pt}
		\caption{The comparison of existing spatial domain registration and the proposed spectral domain registration.  }
		\label{CF0}
		\vspace{-5pt}
	\end{figure}

	To address these issues, we propose solving the HSI-MSI registration problem from the spectral domain.  Specifically,  a lightweight  spectral prior learning (SPL) network is proposed to capture the intrinsic spectral features of the HSIs. Next, we  conduct prior-guided spectral super-resolution on the MSI to generate an initial high spatial resolution HSI (HR-HSI). Then, a registered HSI can be obtained after spatially downsampling the HR-HSI, with the option to manually determine the downsampling process, as illustrated in Figure \ref{CF0}. Moreover, we employ subspace representation method \cite{Hysure} to enhance the robustness of the SPL network, and propose a cyclic training strategy (CTS) to alleviate the spectral  deviation caused by the misregistration.  We refer to this registration approach as spectral domain registration (SDR), which offers the following advantages over spatial domain registration:  1) Despite the significant difference in spatial resolution between the images, the SDR  method can maintain robust  and  exhibit superior performance compared to  spatial domain registration approaches, since the spectral information of the HSI remains well preserved. 2) By tackling registration in spectral domain, the SDR approach enables processing at a substantially reduced scale,  thereby significantly decreasing computational time. 3) Through spectral-domain registration, the proposed SDR method can effectively preserve the spectral integrity of HSIs throughout the registration process.

\begin{figure}[t]
	\vspace{-2pt}
	\centering
	\!\!\includegraphics[width=9.2cm]{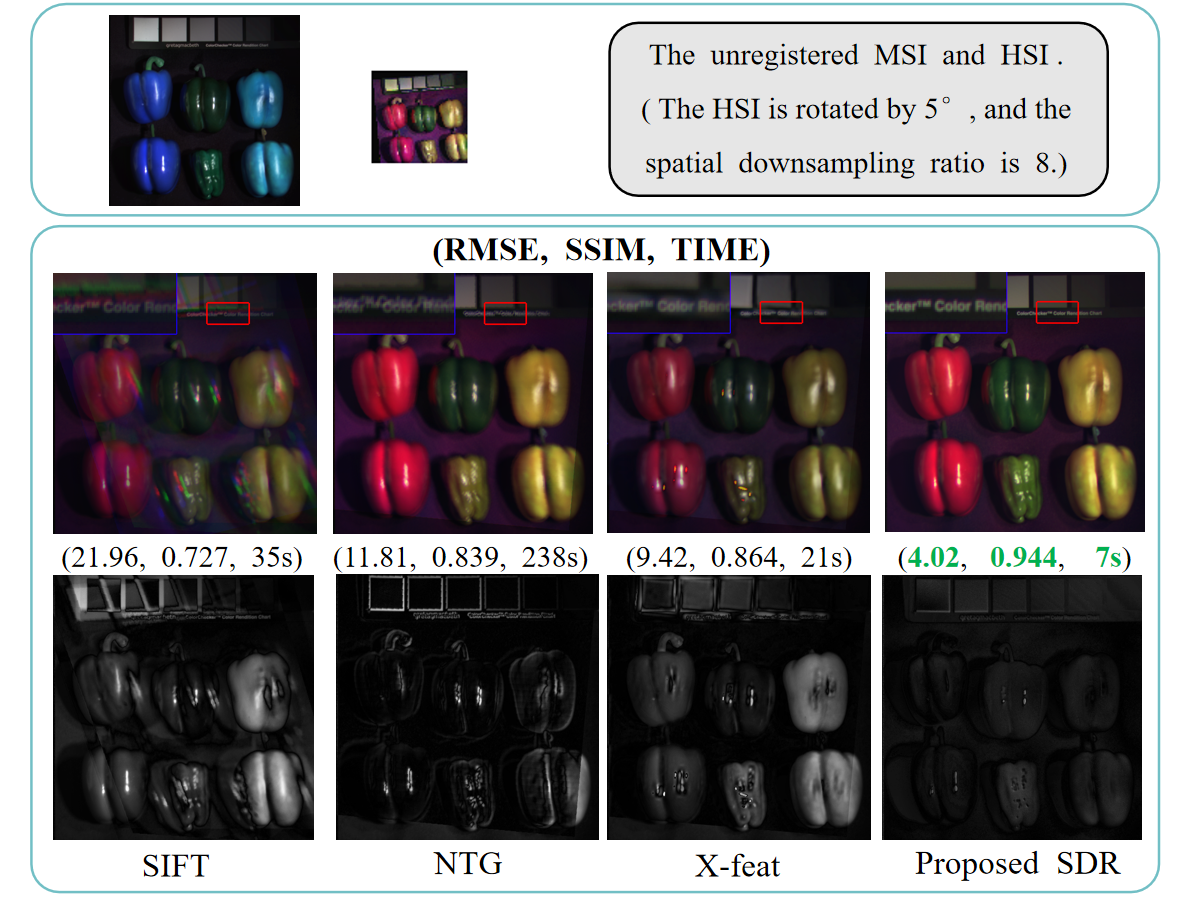}
	
	\vspace{-2pt}
	\caption{Image registration results on the `Peppers' dataset using SIFT \cite{SIFT}, NTG \cite{NTG}, X-feat \cite{Xfeat} and the proposed SDR. The second row presents the blended images of the registered result and the ground truth, while the third row shows the corresponding difference image. }
	\label{CF00}
	\vspace{-4pt}
\end{figure}

	These advantages can be supported by Figure \ref{CF00}, which compares the proposed SDR approach with several well-known registration methods. The results confirm that the SDR method significantly outperforms the other methods in terms of registration performance under a spatial downsampling ratio of 8, where one pixel in the HSI corresponds to 64 pixels in the MSI. In this case,  the SDR method achieves an RMSE that is 5.4 units lower than that of X-feat. Moreover, the SDR method  requires considerably less computational time. 
	
	Subsequently, we propose a subspace representation-based (SR-based)  blind sparse fusion (BSF) model to fuse the   registered HSI-MSI pairs. \textit{Notably, we only need to estimate the spectral degradation operator since the spatial degradation operator  is determined manually.} This reduces the burden of the blind fusion task. To avoid error accumulation, the BSF model performs the fusion and degradation operator estimation tasks simultaneously. The SR-based approaches are generally sensitive to the choice of subspace dimensionality. To address this issue, the proposed BSF model leverages group sparsity to effectively characterize the low-rank structure along the spectral dimension of HSIs. This strategy not only reduces the algorithm's dependence on accurate dimensionality selection, but also avoids the computationally expensive singular value decomposition  required by nuclear norm-based regularization methods \cite{Yang2025, LTMR}. Then,  we employ the Proximal Alternating Optimization (PAO) algorithm  to solve the BSF model and provide its detailed  convergence analysis.

	The summary of our contributions is outlined as follows:
	\begin{itemize}
		\item  We propose a novel spectral domain registration (SDR) method. Specifically,  a lightweight  SPL network is developed, which  generates a registered HSI by performing spectral super-resolution on the MSI.  In this process, we employ the subspace representation (SR) approach to enhance the robustness of the SDR method. Moreover, a cyclic training strategy (CTS) is proposed to further alleviate spectral distortion in the  registered HSI. 
		
		\item  
		We propose the BSF model for the blind fusion task, which only requires estimating the spectral degradation operator.  Furthermore, the  BSF model utilize group sparsity   to equivalently characterize the low-rankness, which not only reduces the algorithm's sensitivity to rank selection but also decreases the computational complexity.
		
		\item  We solve the BSF  model using PAO algorithm and provide its detailed  convergence analysis. Extensive numerical experiments are conducted on  simulated and real-world HSI datasets to verify the efficacy of our proposed method in registration, fusion, and enhancing classification accuracy. Moreover, our approach requires considerably less time than competing methods.
	\end{itemize}
	
	The remaining sections of this paper are organized as follows. Related work  is  introduced in section \ref{A}.  In section \ref{B}, we propose the   SDR-BSF method.
	Then in section \ref{NN}, numerical experiments on HSI-MSI registration and fusion are presented. Finally, we  make a conclusion in section \ref{G}.

\section{Related work} \label{A}
	
	
	\subsection{HSI-MSI fusion}
	\vspace{-2pt}
	Existing HSI-MSI fusion approaches can primarily be categorized into three types:  matrix factorization (MF)-based, tensor factorization (TF)-based  and deep learning-based methods \cite{IR}.   
	MF-based methods explore the internal correlations within images to reduce problem dimensions or perform component separation. Yokoya \textit{et al.} \cite{CNMF} proposed a coupled nonnegative matrix factorization fusion method based on a linear spectral mixture model.
	The work in \cite{Hysure} proposed a subspace representation method  that exploits the strong correlations within the spectral dimensions of HSI to achieve significant dimensionality reduction of the fusion model. 
	Dong \textit{et al.} \cite{NSSR} developed a Nonnegative Structured Sparse Representation (NSSR) method which jointly estimates the  dictionary and the sparse coefficients.
	
	Recently, TF-based methods have shown superior fusion performance by leveraging the multilinear features of the data \cite{Zhao2}. The target HSI was restructured as a CP approximation in work \cite{CP1}, and an alternating least squares algorithm called Super-resolution TEnsor REconstruction (STEREO) was developed. 
	Zeng \textit{et al.} \cite{CP2024} utilized  CP factorization to characterize the low-rank structure of HSIs with a sparse Bayesian framework for adaptive rank determination. 
	Li \textit{et al.} \cite{CSTF} proposed a methodology employing coupled sparse tucker factorization (CSTF) for HSI-MSI fusion. The work in \cite{LXL} integrated linear unmixing with low Tensor-Train decomposition into the fusion model, incorporating a sparse prior to enhance the fusion process.  Xu \textit{et al.} \cite{XY1} employed a tensor-product (t-product)  to decompose the target HSI and utilized non-local clustering sparse representation method. 
	
	Deep learning-based methods employ neural networks to extract intrinsic features from HSI and MSI data, then synthesizing a fused image that effectively integrates the spatial details and spectral characteristics from both datasets \cite{Gao2023} \cite{Liu2025}. Palsson \textit{et al.} \cite{3DCNN} used a 3D convolutional neural network to fuse the HSI and MSI, while reduced
	the dimension of  HSI by using principal component analysis.  Dian \textit{et al.} \cite{CNND} incorporated a subspace representation method into a plug-and-play framework, in which a CNN-based denoiser is integrated into the ADMM algorithm as the proximity operator. Xie \textit{et al.} \cite{Xie2022a} proposed an interpretable fusion network by unfolding the proximal gradient algorithm to solve the  model, which facilitates an easy intuitive observation and analysis on what happens inside the network.  Wang \textit{et al.} \cite{Wang2023} developed a spatial-spectral implicit neural representation fusion network to adequately restore the continuous spatial and spectral information of the images.  Liu \textit{et al.} \cite{Liu2024} proposed a low-rank Transformer network (LRTN) for HSI-MSI fusion, exploiting both the intrinsic correlations and the global spatial correlations within images.
	
	The aforementioned fusion methods generally assume that the spatial and spectral degradation operators are known. In practice, however, these operators may be unknown, which is referred to as blind fusion \cite{CP1}.
	Several alternating optimization algorithms have been proposed to specifically estimate the degradation operators under accurate registration conditions \cite{ZSL,Hysure}. To prevent error accumulation, Yang \textit{et al.} \cite{Yang2025} incorporated triple decomposition and  the estimation of the degradation operators simultaneously with the fusion task. Moreover, some approaches leverage the powerful  nonlinear fitting capabilities of neural networks to design modules that approximate the spatial and spectral degradation processes \cite{ UDA, Zhang2024}. Wang \textit{et al.} \cite{Wang2024} developed an unsupervised blind fusion method based on Tucker decomposition and spatial–spectral
	manifold learning.
	

	\subsection{Registration-fusion method}
	Due to the fact that HSIs and MSIs are typically captured by different sensors, the image pairs obtained in practice are often unregistered, which severely affects the subsequent fusion process. This has prompted researchers to jointly address the challenges of registration and fusion.
	In the case where the degradation operators are known, Fu \textit{et al.} \cite{Fu2020} presented an effective approach for simultaneous HSI super-resolution and geometric alignment of the image pair, which incorporates spectral dictionary learning and sparse coding methods. Ying \textit{et al.} \cite{NED} introduced a new registration method named NED for clear–blur image similarity measurement and incorporated interpolation into the fusion process.  For cases where the degradation operators are unknown, the task of registration and fusion becomes even more challenging. Recently, some deep learning-based methods have been proposed. Qu \textit{et al.} \cite{Qu2022} developed an unregistered and unsupervised mutual Dirichlet-Net (u2-MDN), which employed  a collaborative $l_{2,1}$ norm  as the reconstruction error. Zheng \textit{et al.} \cite{Zheng2022} embed the linear spectral unmixing method into the fusion network and introduced the STN \cite{STN} as the registration module within the network. 
	Guo \textit{et al.} \cite{Guo2023} developed  a stereo cross-attention network  based on a Transformer to extract the abstract features of the unregistered images, then reconstruct to obtain the fused image.  In \cite{Qu2024}, a multi-scale registration-fusion consistency physical perception model (RFCM) is developed, which leverages optimization algorithms  to guide the fusion process.
	
    \section{The proposed SDR-BSF method}\label{B}

	The HSIs and MSIs can be considered as 3-D tensors. We represent the target HR-HSI as $\mathcal{X}\in \mathbb{R}^{M\times N\times H}$, where $M$ and $N$ denote the two spatial dimensions, and $H$ represents the number of spectral bands. Similarly, the obtained HSIs and MSIs are denoted as $\mathcal{Y}\in \mathbb{R}^{m\times n\times H}$ and $\mathcal{Z}\in \mathbb{R}^{M\times N\times h}$, respectively, where $m< M$, $n< N$, and $h< H$. We denote $\textbf{X}\in \mathbb{R}^{H\times MN}$, $\textbf{Y}\in\mathbb{R}^{H\times mn}$, and $\textbf{Z}\in\mathbb{R}^{h\times MN}$ as the mode-3 unfolding matrices \cite{Kolda} of $\mathcal{X}$, $\mathcal{Y}$ and $\mathcal{Z}$, respectively.
	
	The derived HSI $\textbf{Y}$ can be perceived as a spatially degraded variant of the target HR-HSI $\textbf{X}$.
	Therefore, the relationship between $\textbf{Y}$ and $\textbf{X}$ can be expressed as:
	\begin{equation}\label{C1}
		\begin{array}{l}
			\textbf{Y}=\textbf{X}\textbf{BS}+\textbf{N}_1,
		\end{array}
	\end{equation} 
	where $\textbf{B}\in \mathbb{R}^{MN\times MN}$ is a spatial blurring matrix representing the hyperspectral sensor’s point spread function (PSF),  which is assumed band-independent and operates under circular boundary conditions \cite{Hysure}, $\textbf{S}\in \mathbb{R}^{MN\times mn}$ is a downsampling matrix. Matrix $\textbf{N}_1$ represents the noise, with the assumption that it follows the Gaussian distribution.
	Similarly, the  MSI $\textbf{Z}$ can be considered as the  degraded version of  $\textbf{X}$ along the spectral mode. The relationship  can be expressed as:
	\begin{equation}\label{C2}
		\begin{array}{l}
			\textbf{Z}=\textbf{R}\textbf{X}+\textbf{N}_2,
		\end{array}
	\end{equation} 
	where $\textbf{R}\in \mathbb{R}^{h\times H}$ is the spectral response matrix and depends on the spectral response function (SRF) of sensor, $\textbf{N}_2$ denotes the Gaussian noise contained in the observed MSI. 
	
	Equations (\ref{C1}) and (\ref{C2}) may not hold true simultaneously since  \textbf{Y} and \textbf{Z} might not be properly registered. 
	The existing registration process typically treats the MSI $\mathcal{Z}$ as the reference and applies a spatial transformation $\tau$ to the HSI $\mathcal{Y}$. The registered HSI, denoted as $\mathcal{Y}_R$, is then defined as: 
	\begin{equation}
		\mathcal{Y}_{R(i)} := \mathcal{Y}_{(i)}(\tau[x,y]), \quad i = 1,2,\dots,H,
	\end{equation}  
	where $\mathcal{Y}_{(i)}$ denotes the $i$-th spectral band.
	However, due to factors such as significant differences in image spatial resolution,  most image registration methods generally cannot achieve satisfactory accuracy \cite{Ren2023}. Moreover, the process of spatial registration  is frequently time-consuming  and may potentially introduce deviations in the spectral dimension of the HSIs. Therefore, we consider whether the registration problem can be addressed  from the spectral domain.
	From this point, we propose to fix the HSI and transform the MSI,  allowing the HSI to provide spectral prior information for MSI spectral super-resolution. In this way, we can obtain a preliminary HR-HSI.  After spatial downsampling, we obtain a new HSI that, theoretically, is precisely registered with the MSI.
	
	\textbf{The first challenge is: how to leverage the  spectral information from the HSI to guide  spectral super-resolution on the MSI}. Our approach is to learn a mapping that transforms the MSI into HSI. This mapping, denoted as $\mathcal{F}: \mathbb{R}^{n_1 \times n_2 \times h} \to \mathbb{R}^{n_1 \times n_2 \times H}$, functionally serves as the inverse of the spectral response operator \textbf{R} in  equation (\ref{C2}). However,  \textbf{R} is a matrix with far more columns than rows, rendering it devoid of a true inverse matrix. 
	Therefore, we use a neural network to approximate the mapping \(\mathcal{F}\), which is referred to as the spectral prior learning (SPL) network. 
	Once trained, the SPL network will be employed to perform spectral super-resolution on the MSI $\mathcal{Z}$, yielding an initial HR-HSI. By spatially downsampling the HR-HSI, we can obtain  a registered HSI $\mathcal{Y}_R$. Finally, we fuse $\mathcal{Y}_R$ and  $\mathcal{Z}$ using the BSF model. The specific workflow is as follows:
	\begin{figure}[t]
		\vspace{-6pt}
		\centering
		\!\!\!\!\includegraphics[width=8.5cm]{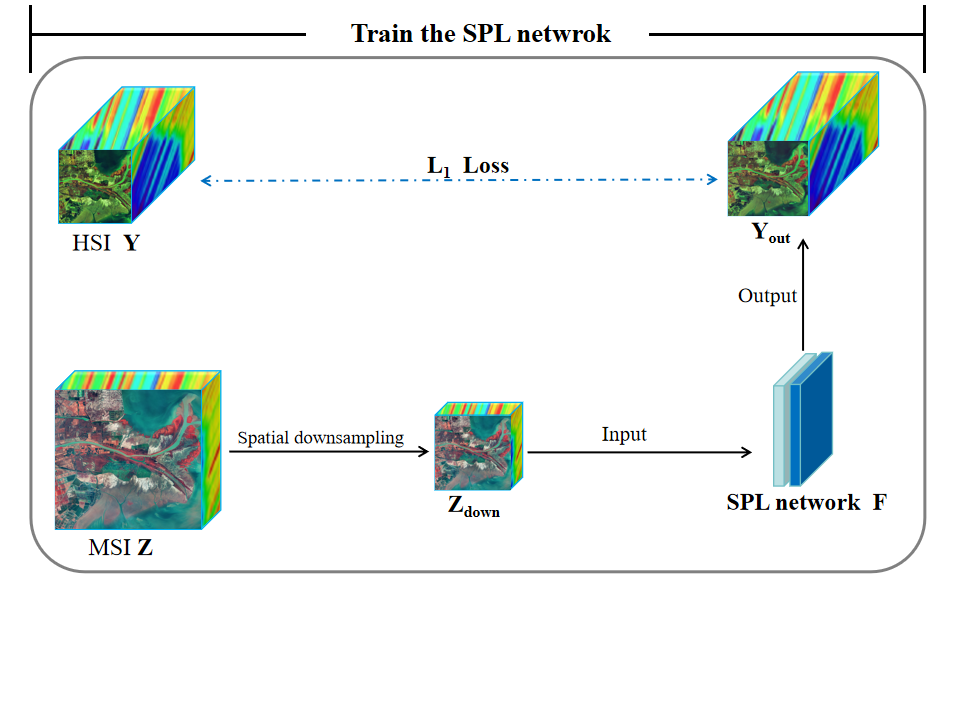}
		
		\vspace{-36pt}
		\caption{ The procedure of  training the SPL network.   }
		\label{CF2}
		\vspace{-3pt}
	\end{figure}

	
	\textbf{Step 1: Train the SPL network.}  Firstly, we spatially downsample the MSI to align its spatial dimensions with those of the HSI. Using a Python command, we denote:
	\begin{equation}
		\mathcal{Z}_{down}:=\mathcal{Z}[1:M:d,~ 1:N:d, ~:],
	\end{equation}
	where $d$ represents the downsampling ratio.
	Then, we input  $\mathcal{Z}_{down}$ into the network $\mathcal{F}$ for spectral super-resolution, yielding another HSI, denoted as $\mathcal{Y}_{out}$, i.e., $\mathcal{Y}_{out}=\mathcal{F}(\mathcal{Z}_{down})$. The loss function is defined as:
	\begin{equation}
		Loss:= \Vert \mathcal{Y}_{out}-\mathcal{Y} \Vert_{l_1}.
	\end{equation}
	where $\Vert \cdot \Vert_{l_1}$ is the $l_1$ norm. The procedure of training the SPL network is illustrated in Figure \ref{CF2}.
	Our objective is to ensure that $\mathcal{Y}_{out}$ is as spectrally similar to $\mathcal{Y}$ as possible. Therefore, we employ  spatial cropping technique  to highlight the spectral dimension information while increasing the amount of training data. Finally, we optimize the network parameters using adaptive moment estimation (Adam) \cite{ADAM} algorithm. 
	
	\textbf{Step2: Obtain registered HSIs.} By inputting the original MSI $\mathcal{Z}$ into the trained network $\mathcal{F}$, we can obtain an initial HR-HSI $\mathcal{F}(\mathcal{Z})$. Subsequently, we spatially downsample $\mathcal{F}(\mathcal{Z})$ to obtain a registered HSI, which is denoted as:
	\begin{equation}\label{C4}
		\mathcal{Y}_{R}:=\textbf{Fold}_3((\mathcal{F}(\mathcal{Z})_{(3)})\widehat{\textbf{B}}\textbf{S}),
	\end{equation}
	where $(\cdot)_{(3)}$ represents the mode-3 unfolding matrix of a tensor and $\textbf{Fold}_3(\cdot)$ denotes its inverse operator. 
	Here, we can choose the blur kernel $\widehat{\textbf{B}}$ manually. Typically, we  choose the  kernel $\widehat{\textbf{B}}$ that is much stronger than the original \textbf{B}, so that the effect of \textbf{B} becomes negligible. Then, the obtained $\mathcal{Y}_{R}$ is spatially registered  with the MSI $\mathcal{Z}$ since it comes from $\mathcal{F}(\mathcal{Z})$.
	
	\textbf{Step3: Blind fusion.} We utilize the BSF model to perform blind fusion on $\mathcal{Y}_R$ and $\mathcal{Z}$, which do not need to estimate the blur kernel since we have pre-set $\widehat{\textbf{B}}$  in  (\ref{C4}). \qed

\begin{figure}[t]
		\vspace{-5pt}
		\centering
		\!\!\!\!\!\!\!\includegraphics[width=8.5cm]{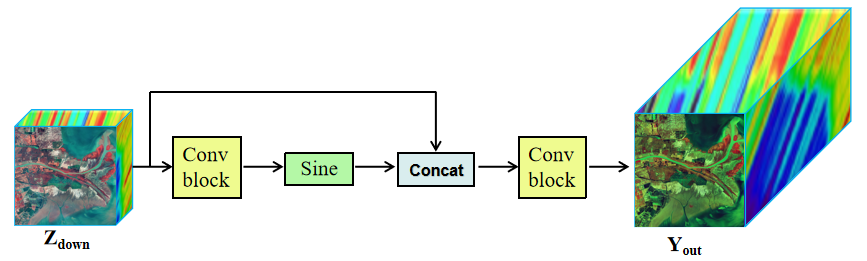}
		
		\vspace{-7pt}
		\caption{ The architecture of the SPL Network.   }
		\label{CF3}
		\vspace{-7pt}
	\end{figure}

	The SPL network $\mathcal{F}$  comprises a residual network constructed with two layers of convolutions, as shown in Figure \ref{CF3}. The activation function Sine \cite{Sine} is applied in our network to enhance its nonlinear fitting capability.  

	
	\subsection{Improving the spectral domain registration (SDR) process}
	Since the initial HSI \(\mathcal{Y}\) and MSI \(\mathcal{Z}\) are not registered, the above SDR process may introduce spectral  deviations, resulting in inaccuracies in the spectral information of the obtained \(\mathcal{Y}_R\). Moreover, the spectral super-resolution process lacks sufficient robustness to noise.
	Therefore, \textbf{the second challenge is: how to improve the accuracy and robustness of spectral domain registration process.} 
	
	\subsubsection{Enhance the robustness of SDR}
	HSIs typically exhibit strong spectral correlations. \textit{Considering that $\mathcal{Y}_R$ primarily inherits the spectral information from $\mathcal{Y}$, they should approximately lie in the same spectral subspace}. Therefore,  we leverage subspace representation (SR) method \cite{Hysure}, which projects the HSI $\mathcal{Y}$ into a lower-dimensional subspace:
	\begin{equation}
		\begin{array}{l}
			\mathcal{Y}=\mathcal{A}\times_3\textbf{D},
		\end{array}
	\end{equation} 
	where $\mathcal{A}\in \mathbb{R}^{M\times N\times L}$  is  the coefficient tensor,  $\textbf{D}\in \mathbb{R}^{H\times L}$ denotes the dictionary and $L$ represents the dimension of the subspace spanned by the columns of $\textbf{D}$. The `$\times_3$' represents the mode-3 tensor-matrix product  \cite{Kolda}.
	Regarding the selection of the dictionary \textbf{D}, we employ the classical Truncated-SVD method, which is outlined as follows:
	\begin{equation}\label{Dic}
		\begin{array}{l}
			[\textbf{U},\Sigma,\textbf{V}]=\text{svd}(\textbf{Y})\quad \text{and}  \quad \textbf{D}=\textbf{U}(: , 1:L),
		\end{array}
	\end{equation}
	where  $\Sigma$ denotes the matrix whose diagonal elements are the singular values of $\textbf{Y}$, $\textbf{U}$ and $\textbf{V}$ are the left and right singular matrices, respectively. Then, we project  $\mathcal{Y}_{out}$ into the subspace, which can be expressed as:
	\begin{equation}
		\begin{array}{l}
			\mathcal{Y}_{out}=\mathcal{A}_{out}\times_3\textbf{D},
		\end{array}
	\end{equation} 
	where $\mathcal{A}_{out}\in \mathbb{R}^{M\times N\times L}$  is  the coefficient tensor. In this way, we only need to learn the mapping from MSI to the coefficient tensor, denoted as $\mathcal{G}:\mathbb{R}^{n_1\times n_2 \times h}\to \mathbb{R}^{n_1\times n_2 \times L}$.    The relationship  can be expressed as
	\begin{equation}
		\begin{aligned}
			\mathcal{A}_{out}=\mathcal{G}(\mathcal{Z}_{down}), \quad~
			\mathcal{F}(\cdot)=\mathcal{G}(\cdot)\times_3\textbf{D}.
		\end{aligned}
	\end{equation}
	The loss function  is defined as  $\Vert\mathcal{A}-\mathcal{A}_{out}\Vert_{l_1}$. Since  $L$ is typically  much smaller than $H$, training the network $\mathcal{G}(\cdot)$ is more efficient than directly training  $\mathcal{F}(\cdot)$. Therefore, the SR method can not only enhance the robustness of the model, but also improve the efficiency of spectral super-resolution.

    \subsubsection{Mitigating \(\mathcal{Y}_R\)'s spectral  deviation}
\begin{figure}[t]
		\centering
		\!\!\!\!\!\!\includegraphics[width=9.1cm]{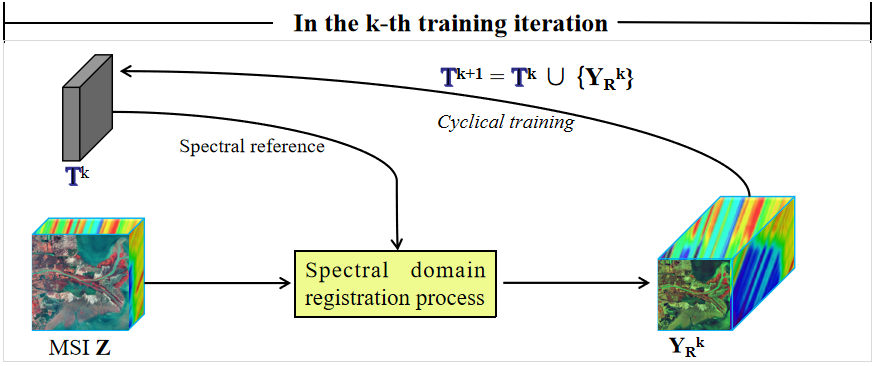}
		
		\vspace{-5pt}
		\caption{The flowchart of the cyclic training strategy (CTS).  }
		\label{CF4}
	\end{figure}

	In fact, the original HSI $\mathcal{Y}$ possesses accurate spectral information but is spatially misregistered with respect to the MSI $\mathcal{Z}$. Conversely, the $\mathcal{Y}_R$ we obtain is spatially  registered with the MSI $\mathcal{Z}$, but its spectral information is potentially biased. \textit{Both  possess part of the desired properties.}
	Therefore, we propose a cyclic training strategy (CTS), which involves incorporating the generated $\mathcal{Y}_R$ from each iteration into the training set,  allowing the network to undergo repeated training.   
	Specifically, we denote the training set as $\mathbb{T}$. Initially, we define $\mathbb{T}^0=\{\mathcal{Y}\}$. At the end of each training iteration, we incorporate the obtained $\mathcal{Y}_R$ into the training set $\mathbb{T}$, i.e.,
	\begin{equation}\label{D10}
		\mathbb{T}^{k+1}=\mathbb{T}^{k}\cup \{\mathcal{Y}_R^{(k)}\},~ k=0,1,2,...,
	\end{equation}
	where $\mathcal{Y}_R^{(k)}$ represents the output of the $k$-th training iteration, as shown in Figure \ref{CF4}.  Through this iterative training strategy, we can obtain a sufficient amount of training data to train a more stable and higher-quality network. As a  result, the final obtained $\mathcal{Y}_{R}$ not only maintains spatial registration but also exhibits significantly enhanced spectral accuracy. 
	
	\vspace{6pt}
	Finally, by integrating \textit{1)} and \textit{2)}, we project each element in $\mathbb{T}$ into the subspace spanned by \textbf{D} in each training iteration. Therefore, the actual training set in the $k$-th training iteration is:
	\begin{equation}\label{D11}
		\mathbb{T}_{\textbf{D}}^k:=\{\mathcal{Y}\times_3 \textbf{D}^\top ~|~ \mathcal{Y}\in \mathbb{T}^k\},
	\end{equation}
	and we minimize the $l_1$ loss between $\mathcal{A}_{out}$ and $\mathbb{T}_{\textbf{D}}^k$.
	The entire training process is summarized in Algorithm \ref{AL1}.
	\begin{algorithm}
		\caption{The procedure of the SDR method.}
		\parskip 0.3mm
		\hspace*{0.02in} {\bf Input:} $\mathcal{Y}$, $\mathcal{Z}$, \textbf{D}.
		
		\hspace*{0.02in} \textbf{Initialize:} $\mathbb{T}^0=\{\mathcal{Y}\}$, $\mathcal{Z}_{down}$. Set $k=0$.
		
		\quad1: Obtain $\mathcal{A}_{out}$: $\mathcal{A}_{out}=\mathcal{G}(\mathcal{Z}_{down})$;
		
		\quad2: Update the set $\mathbb{T}_{\textbf{D}}^k$ by (\ref{D11});
		
		\quad3: Train the SPL network $\mathcal{G}$ to minimize the $l_1$ loss between
		
		\qquad $\mathcal{A}_{out}$ and the set $\mathbb{T}_\textbf{D}^k$;
		
		\quad4: Obtain the HR-HSI, i.e., $\mathcal{F}(\mathcal{Z})=\mathcal{G}(\mathcal{Z})\times_3\textbf{D}$;
		
		\quad5: Obtain $\mathcal{Y}_R^{(k)}$ by  downsampling $\mathcal{F}(\mathcal{Z})$ as described in (\ref{C4});
		
		\quad6: Update set $\mathbb{T}^{k+1}$ as (\ref{D10});
		
		\quad7: If  the maximum iteration count has not been reached, 
		
		\qquad  set $k = k+1$  and return to step1;
		
		\hspace*{0.02in} {\bf Output:}  $\mathcal{Y}_R^{k}$.
		\label{AL1}
	\end{algorithm}

	\begin{remark}
		When dealing with significant unregistered scale discrepancies, performing a  coarse pre-registration of the images prior to applying the SDR method can generally enhance its registration performance. This approach is particularly applicable to remote sensing geo-images that allow pre-registration based on geographic coordinates.
	\end{remark}
	
	
	\subsection{The blind sparse fusion (BSF) method}
	Next, we  propose the BSF model to obtain the final fused image. Unless otherwise specified, \textbf{Y}, \textbf{Z} and \textbf{X} will represent the mode-3 unfolding matrices of $\mathcal{Y}_R$, $\mathcal{Z}$ and $\mathcal{X}$, respectively.
	
	Considering the strong correlation between different spectral bands of HSI, we leverage subspace representation method for \textbf{X}, i.e.,
	\begin{equation}\label{13}
		\textbf{X} = \textbf{DA},
	\end{equation}
	where $\textbf{D}\in \mathbb{R}^{H\times r}$ and $\textbf{A}\in \mathbb{R}^{r\times MN}$. The matirx \textbf{D} is directly defined  as  equation (\ref{Dic}).   However,  the fusion methods based on (\ref{13}) are typically sensitive to the choice of $r$. Some approach involves utilizing the nuclear norm-based regularizer to promote low-rank properties \cite{Yang2025, LTMR}, which leads to high-quality reconstructions without the need to estimate $r$.  Nevertheless, solving the subproblem requires performing SVD repeatedly, which is computationally expensive.	To address these issues, we utilize the following theorem.
	\begin{theorem}\cite{Li2024}\label{rank}
		For any matrix $\textbf{X}\in \mathbb{R}^{m\times n}$ with rank(\textbf{X}) $\le r$, we have
		\begin{equation}
			\begin{aligned}
				\text{rank}(\textbf{X})&=\min\limits_{\textbf{X}=\textbf{D}\textbf{A}} \Vert \textbf{D}^\top \Vert_{2,0}=\min\limits_{\textbf{X}=\textbf{D}\textbf{A}} \Vert \textbf{A} \Vert_{2,0},
			\end{aligned}
		\end{equation}
		with $\textbf{D}\in \mathbb{R}^{m\times r}$ and  $\textbf{A}\in \mathbb{R}^{r\times n}$.
		Here $\Vert \textbf{A} \Vert_{2,0}:=\sum_{i}^{}\Vert\textbf{a}_i\Vert^0_F$ (adopting the convenience that $0^0=0$) denotes  the number of non-zero rows of \textbf{A} ($\textbf{a}_i$ represents the $i$th row of \textbf{A} and $\Vert \cdot \Vert_F$ represents the Frobenius norm).
	\end{theorem}
	
	Theorem \ref{rank} implies that \textit{even if the chosen $r$ is relatively large, we can leverage group sparsity regularization to promote the low-rank property of the target $\textbf{X}$, thereby reducing the dependence on the selection of parameter $r$.}
	Therefore, we  propose the blind sparse fusion (BSF) model as follows:
	\begin{equation}\label{BSF}
		\begin{aligned}
			&\min\limits_{\textbf{A},\textbf{R}} \Vert \textbf{DABS}-\textbf{Y} \Vert_F^2+\Vert \textbf{RDA}-\textbf{Z} \Vert_F^2 +\alpha\Vert \textbf{A} \Vert_{2,\psi},\\
			&\enspace \text{s.t.}\quad  \textbf{R}\ge 0,
		\end{aligned}
	\end{equation}
	where $\alpha$  is the weight parameter, and  $\Vert \textbf{A} \Vert_{2,\psi}:=\sum_{i}^{}\psi(\Vert\textbf{a}_i\Vert)$ is the non-convex relaxation of $\Vert \textbf{A} \Vert_{2,0}$. The function $\psi$ can be chosen as   the CapL1 function, which is defined as $\psi_\rho(x)=\min\{1,\frac{x}{\rho}\}$ for a fixed parameter $\rho>0$.
	According to \cite{Pan2021}, the proximal mapping of  $\Vert \cdot \Vert_{2,\psi}$ is
	\begin{equation}\label{prox}
		\text{prox}_{\lambda\psi}(x)=
		\begin{cases} 
			(\|x\|-\frac{\lambda}{\rho})_+\dfrac{x}{\|x\|},   & \text{if } \|x\|\le \rho+\frac{\lambda}{2\rho},\\
			x, & \text{otherwise},
		\end{cases}
	\end{equation}
	where $(\cdot)_+:=\max\{\cdot,~0\}$. The complexity of computing (\ref{prox}) is much lower than that of computing the  proximal operator of the nuclear norm \cite{RPCA2011}.
	Note that we only need to estimate the spectral degradation operator \textbf{R}, which avoids the BSF model becoming more ill-posed. 
	
	To solve the BSF model, we employ the  Proximal Alternating Optimization (PAO) algorithm. For convenience, we denote the objective function of the BSF model as $g(\textbf{A},\textbf{R})$. In the $k$-th iteration, we  add proximal terms to the objective function for each variable.  The iterative scheme of PAO is
	\begin{align}
			\textbf{A}^{k+1}&={\rm arg}\min\limits_{\textbf{A}}~ g(\textbf{A},\textbf{R}^k)+\dfrac{\lambda}{2}\Vert \textbf{A}-\textbf{A}^k \Vert_F^2, \label{tt1}\\
			\textbf{R}^{k+1}&={\rm arg}\min\limits_{\textbf{R}\ge 0}~ g(\textbf{A}^{k+1},\textbf{R})+\dfrac{\lambda}{2}\Vert \textbf{R}-\textbf{R}^k \Vert_F^2, \label{tt2}
	\end{align}
	where  $\lambda>0$ denotes the proximal  parameter. 
	We  update variables $\textbf{A}^{k+1}$ and $\textbf{R}^{k+1}$ alternatively.  The PAO algorithm for solving the BSF model is summarized  in Algorithm \ref{PAO}.

    \begin{algorithm}
				\caption{ PAO algorithm for solving the BSF model}
				\parskip 0.8mm
				\hspace*{0.02in} {\bf Input:} \textbf{Y}, \textbf{Z}, $\alpha$,   $\rho$ and $\lambda$. 
				
				\hspace*{0.02in} \textbf{Initialize:} $\textbf{A}^k$ and $\textbf{R}^k$ . Set $k=0$.
				
				\quad1: Update $\textbf{A}^{k+1}$  by  solving the subproblem (\ref{tt1});
				
				\quad2: Update $\textbf{R}^{k+1}$  by  solving the subproblem (\ref{tt2});
				
				\quad3: If the termination criterion is not met,
				
				\qquad  set $k = k+1$  and return to step1; 
				
				\hspace*{0.02in} {\bf Output:}  $\textbf{X}=\textbf{DA}^{k+1}$.
				\label{PAO}
			\end{algorithm}
	Next, we present the convergence analysis of the algorithm.

	\begin{theorem}\label{T2} Let $\{\mathcal{P}^{k}=(\textbf{A}^k,\textbf{R}^k)\}_{k\in\mathbb{N}}$ be the sequence generated by Algorithm \ref{PAO}. If  $\{\mathcal{P}^{k}\}_{k\in\mathbb{N}}$ is bounded, then $\{\mathcal{P}^{k}\}_{k\in\mathbb{N}}$ converges to a critical point of the objective function of model (\ref{BSF}) and  
		$$
		\sum_{k=0}^{+\infty}||\mathcal{P}^{k+1}-\mathcal{P}^{k}|| < +\infty.
		$$
	\end{theorem}

    \vspace{0pt}
	The detailed  process of the algorithm and the convergence analysis are included in the supplementary materials.
	
	From the conclusion of the Theorem \ref{T2}, we can deduce that  ${\rm lim}_{k\to +\infty}\Vert \mathcal{P}^{k+1}-\mathcal{P}^k \Vert=0$. Therefore, the relative change between two consecutive steps can be directly employed  as the termination criterion for the PAO algorithm.

	
	\section{NUMERICAL EXPERIMENTS}\label{NN}
	In this section, we  evaluate the proposed SDR-BSF method through a series of experiments conducted on publicly available HSI datasets.  Initially, to assess the registration performance of the SDR method, we compare it with: the feature-based SIFT \cite{SIFT}, intensity-based  NTG \cite{NTG}, and deep learning-based  X-feat \cite{Xfeat} approaches. Meanwhile, two blind fusion algorithms, TriD \cite{Yang2025} and ZSL \cite{ZSL}, are selected to fuse aligned image pairs to  validate the overall effectiveness of the proposed SDR-BSF method. Next, we compare the SDR-BSF method with several well-known  fusion methods on both simulated and real datasets. These methods include  model-driven approaches: Hysure \cite{Hysure}, Integrated \cite{Zhou2020}, and NED \cite{NED}, as well as  deep learning-based methods: NonregSR \cite{Zheng2022}, PMIRFCo \cite{PMI}, DFMF \cite{Guo2022}, HPWRL \cite{Nie2024} and IR-ArF \cite{Qu2025}. Finally, ablation studies are  conducted to  validate the effectiveness of each novel component in our approach.
	
	\subsection{Datasets}
	
	\textit{1). Simulated datasets.}
	
	(a). Pavia University and Pavia Center \cite{Pavia}: The Pavia University and Pavia Center datasets, following processing, respectively encompass 103 and 102 spectral bands, each spanning a wavelength spectrum from 0.43µm to 0.86µm.  We select 93 bands from both scenes and crop a region of interest from the up-left corner, resulting in the image size of 256 × 256 × 93. 
	To generate the HSIs,   we apply the spatial downsampling with  scale factor (\textbf{sf}) of 4  and 8 to the Pavia Center and Pavia University, respectively. Additionally, the HR-MSI  is generated using a four-band IKONOS-like reflectance spectral response filter \cite{sy}.
	
	(b).  CAVE$^1$: The CAVE dataset comprises images with 31 bands, spanning the spectral range from 400 nm to 700 nm at 10 nm intervals. All images within the dataset were uniform in dimensions, measuring 512 × 512 × 31.  We selected the `Face', `Superballs' and `Toys'  images for our experiments. Specifically, the spatial dimensions of the `Face'  were downsampled by a scale factor (\textbf{sf}) of 8, whereas those of the `Superballs' image were downsampled by $\textbf{sf}=16$. The `Toys'  was downsampled by $\textbf{sf}=32$. We obtained the MSIs with three bands by the Nikon D700 camera$^2$, i.e., $h = 3$.
	
	\footnotetext[1]{https://www.cs.columbia.edu/CAVE/databases/multispectral/ }
	\footnotetext[2]{https://maxmax.com/spectral\_response.htm}

	\textit{2). Real-world Dataset.}
	
	(c). GF1-GF5 \cite{Guo2023}: In the GF1-GF5 dataset,  the size of HSI is 1161 × 1129 × 150, and the size of MSI is 2322 × 2258 × 4. For the HSI, we select bands 11 to 100 and extract a 1024 × 1024 pixel area from the top-left quadrant, using a sampling ratio of 8, resulting in a HSI with dimensions of 128 × 128 pixels across 90 bands. Concurrently, a 2048 × 2048 pixel segment was taken from the same top-left corner of the MSI, employing a sampling interval of 4, which produced a MSI with dimensions of 512 × 512 pixels and 4 bands.
	
	(d). FR2 \cite{FR}: The Panchromatic (Pan) image and HSI of the full resolution dataset  FR2 have been obtained by extracting a  12km × 12km  portion (2400 × 2400 pixels for Pan image and  400 × 400 × 63  pixels for HSI) from the original 30km × 30km PRISMA acquisition. In our experiments, we spatially downsampled the Pan image and HSI by a factor of 4. This resulted in an HSI with dimensions of 100 × 100 pixels and 63 bands, and a Pan image with a size of 600 × 600 pixels.

	\subsection{Implementation Details}
	Subsequently, we provide a  description of our experimental details and the associated settings. 
	
	\textbf{Unregistered settings.}
	For the unregistered case, to quantitatively compare evaluation metrics, we manually introduce deformations to generate unaligned images for comparison. Specifically, we apply three types of transformations to each band of the HSI, including:
	\vspace{-2pt}
	\begin{itemize}
		\item Scaling: we enlarge each band of the HSI by 1.1 times.
		
		\item Rotation: we rotate each band of the HSI by 2 degrees.
		
		\item Pincushion Distortion: we apply pincushion distortion to each band of the HSI to simulate common distortions at the edges of the camera's field of view (nonlinear). The scale of the distortion is quantified by a distortion coefficient, which we set to  $1\times 10^{-2}$.
	\end{itemize}
	\vspace{-2pt}
	The spatial dimensions of the HSI typically change after undergoing various transformations, and we  crop the images back to their original size.

    In addition to simulation experiments, we also conducted tests on real-world datasets: GF1-GF5 and FR2. The images in the GF1-GF5 dataset are unregistered, while the images from the FR2 dataset have undergone pre-registration. To increase the degree of misregistration, we artificially introduced local pincushion distortions to the HSI in FR2 dataset. Furthermore, both the spatial and spectral degradation operators are unknown for the two datasets.

	\textbf{Parameter settings.}  For the SPL network, the convolution kernel size is typically set within the range of 3 to 9, with zero-padding applied. Generally, the greater the degree of misregistration, the larger the convolution kernel size should be, as this allows the network to capture feature correlation over a wider range. The  subspace dimension $L$ is set to 10. For the BSF model, we set  \( \alpha = 2 \times 10^{-1} \), $\nu = 2$ and \( \lambda = 1 \times 10^{-3} \). The parameter $r$ is set within [2, 6]. 

	
	\textbf{Evaluation metrics.} To  assess  the quality of the reconstructed HSI, we choose  the following  evaluation metrics: peak signal-to-noise ratio (PSNR) \cite{MSE},  structural similarity (SSIM) \cite{MSE}, relative dimensionless global error in synthesis (ERGAS) \cite{Wei2015}, spectral angle mapper  (SAM) \cite{Wei2015} and root mean square error (RMSE) \cite{Wei2015}. We also compared the time consumption of each method. The SPL network is trained on a NVIDIA GeForce RTX 4060 GPU, and we run the BSF model in MATLAB R2023a on a 12th Gen Intel® Core™ i5-12450H processor at 2.00 GHz.
	
	\vspace{-1pt}
	\subsection{The effectiveness of SDR-BSF method}
	\vspace{-1pt}
	In this subsection, we  aim to verify the effectiveness of the proposed SDR-BSF method.
	We initially verify the registration performance of the SDR method. Three well-known registration algorithms: the feature-based SIFT \cite{SIFT}, the intensity-based NTG \cite{NTG} and deep learning-based X-feat \cite{Xfeat}  are selected to perform band-wise registration for comparsion.  Table \ref{DT2} records the evaluation metrics for the registered HSI obtained by compared methods on different datasets.
	
	\begin{table}[hpbt]
		\renewcommand\arraystretch{1.1}
		\caption{ The evaluation metrics of  test image registration methods.} 
		\vspace{-0.05cm}      
		\centering
		\begin{tabular}{p{1.4cm}|p{0.1cm}p{1.1cm}p{1.1cm}p{1.1cm}c}
			\hline
			\textbf{Methods}  & & SSIM & RMSE & SAM& TIME(s) 
			\\ \hline
			\multicolumn{6}{c}{\textbf{ Faces (sf = 8, ~scaling)}} \\
			\hline
			~SIFT \cite{SIFT} &  & 0.912 &  14.03 & 19.856 & 37  \\ 
			~NTG \cite{NTG}&  & 0.924 &  12.55 & 17.056 & 152\\  
			~X-feat  \cite{Xfeat}&  & 0.907 &  14.18 & 20.615 & 23 \\  
			~SDR &  & \textbf{0.963}  & \textbf{3.52} & \textbf{11.140} &  \textbf{8} \\ 
			\hline
			\multicolumn{6}{c}{\textbf{ Toys (sf = 32, ~pincushion)}} \\
			\hline
			~SIFT \cite{SIFT}&  & 0.626 &  20.98 & 16.927 & 33 \\ 
			~NTG  \cite{NTG}&  & 0.759 &  15.35 & 23.323 & 147\\  
			~X-feat \cite{Xfeat}&  & 0.699 &  18.36 & 20.156 & 22 \\  
			~SDR  &  &  \textbf{0.965} & \textbf{3.88} & \textbf{11.695} & \textbf{6} \\ 
			\hline
			\multicolumn{6}{c}{\textbf{ Pavia U  (sf = 4, ~rotation)}} \\
			\hline
			~SIFT \cite{SIFT} &  & 0.737 &  16.84 & 6.544 & 25  \\ 
			~NTG \cite{NTG}&  & 0.912 &  14.75 & 8.276 & 231 \\  
			~X-feat \cite{Xfeat}&  & 0.872 &  12.53 & 6.726 & 42 \\  
			~SDR &  & \textbf{0.968}  & \textbf{6.01} & \textbf{3.205} &  \textbf{6} \\ 
			\hline
			\multicolumn{6}{c}{\textbf{ Pavia C  (sf = 16, ~pincushion)}} \\
			\hline
			~SIFT \cite{SIFT}&  & 0.616 &  18.69 & 16.776 & 23 \\ 
			~NTG  \cite{NTG}&  & 0.776 &  15.64 & 11.378 & 229\\  
			~X-feat \cite{Xfeat} &  & 0.717 &  17.81 & 12.619 & 42 \\  
			~SDR  &  &  \textbf{0.960} & \textbf{6.43} & \textbf{5.087} & \textbf{4} \\ 
			\hline
		\end{tabular}
		\label{DT2}
		\vspace{-5pt}
	\end{table}

\begin{figure}[htbp]
		\centering
		\vspace{-1pt}
		\captionsetup[subfloat]{labelsep=none,format=plain,labelformat=empty}
		\subfloat{\includegraphics[width = 1.93cm]{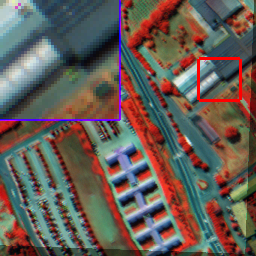}}\!
		\subfloat{\includegraphics[width = 1.93cm]{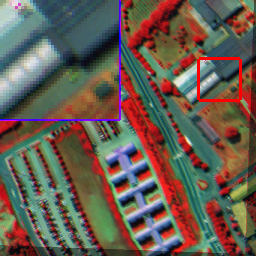}}\!
		\subfloat{\includegraphics[width = 1.93cm]{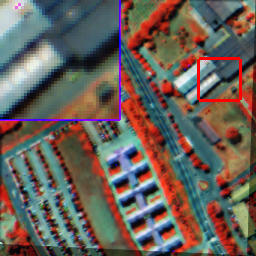}}\!
		\subfloat{\includegraphics[width = 1.93cm]{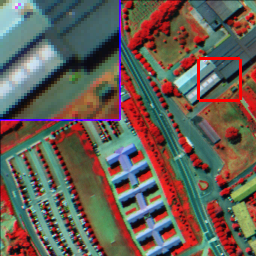}}
		
		\vspace{-9pt}
		
		\subfloat[{\small SIFT}]{\includegraphics[width = 1.93cm]{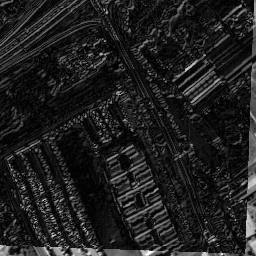}}\!
		\subfloat[{\small NTG}]{\includegraphics[width = 1.93cm]{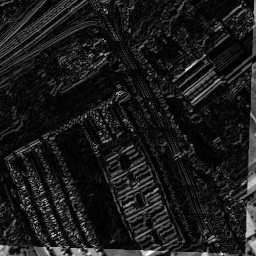}}\!
		\subfloat[{\small X-feat}]{\includegraphics[width = 1.93cm]{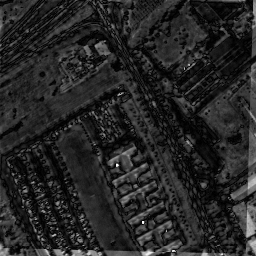}}\!
		\subfloat[{\small SDR}]{\includegraphics[width = 1.93cm]{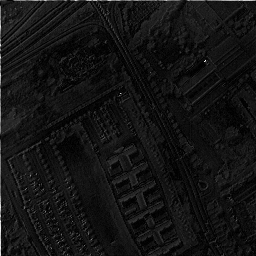}}

		\caption{The registration results  for `Pavia University' (bands 20, 40 and 60,  \textbf{sf} = 4) dataset. The first row shows the blend image of the registered image and the original image, while the second row displays the difference image.\tiny\label{AF0}}
	\end{figure}

	From the table, spatial domain registration approaches tend to exhibit a significant performance degradation with large scale factors. In contrast, the proposed SDR method remains largely unaffected by significant scale factors (sf), which can be primarily attributed to the fact that the spectral information of HSIs is relatively insensitive to substantial sampling variations.
	Furthermore, the entire process of SDR only involves spectral super-resolution, without the need for feature point search and matching as required by SIFT and X-feat, or minimizing a loss function in an iterative algorithm form as with NTG. Therefore, the SDR method  has a significantly lower computational time. Figure \ref{AF0} shows the registration results of each method. As shown in the figure, the registered images produced by the compared methods exhibit more noticeable artifacts and distortions. In contrast, the image registered using the SDR method is highly consistent with the original image. The difference map of SDR is mainly concentrated along the edges, which is  attributed to the smoothing effect of the convolutional neural network. However, since HSIs mainly provide spectral information, the lost texture details can be effectively compensated by the MSIs, and therefore do not compromise the subsequent fusion performance.

\begin{figure*}[htbp]
		\centering
		\vspace{-10pt}
		\captionsetup[subfloat]{labelsep=none,format=plain,labelformat=empty}
		\subfloat{\includegraphics[width = 1.73cm]{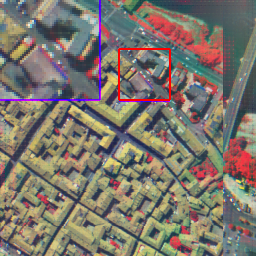}}\!
		\subfloat{\includegraphics[width = 1.73cm]{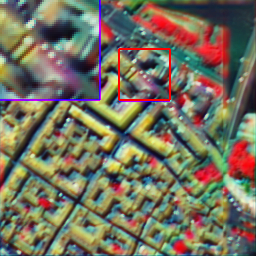}}\!
		\subfloat{\includegraphics[width = 1.73cm]{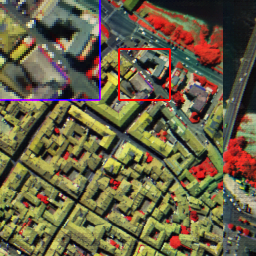}}\!
		\subfloat{\includegraphics[width = 1.73cm]{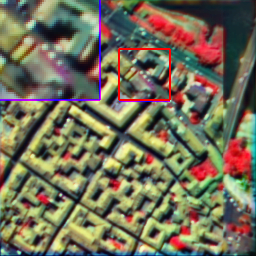}}\!
		\subfloat{\includegraphics[width = 1.73cm]{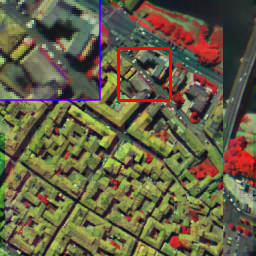}}\!
		\subfloat{\includegraphics[width = 1.73cm]{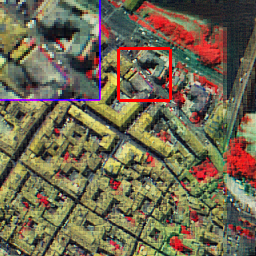}}\!
		\subfloat{\includegraphics[width = 1.73cm]{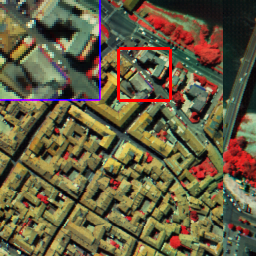}}\!
		\subfloat{\includegraphics[width = 1.73cm]{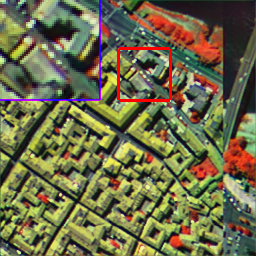}}\!
		\subfloat{\includegraphics[width = 1.73cm]{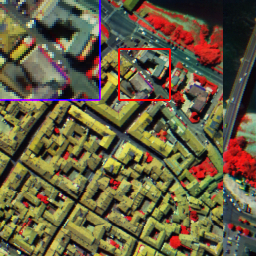}}\!
		\subfloat{\includegraphics[width = 1.73cm]{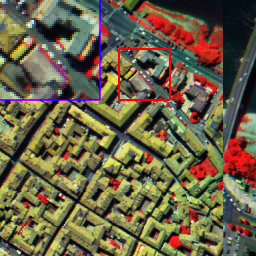}}
		\subfloat{\includegraphics[width = 0.24cm]{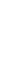}}
		
		\vspace{-9pt}
		
		\subfloat[{\small SIFT-TriD}]{\includegraphics[width = 1.73cm]{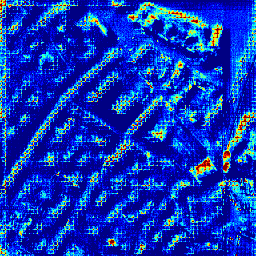}}\!
		\subfloat[{\small SIFT-ZSL}]{\includegraphics[width = 1.73cm]{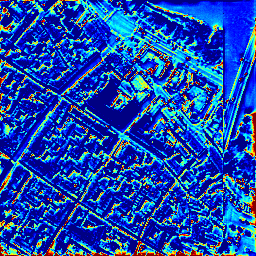}}\!
		\subfloat[{\small NTG-TriD}]{\includegraphics[width = 1.73cm]{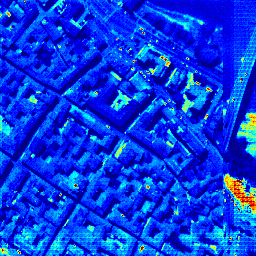}}\!
		\subfloat[{\small NTG-ZSL}]{\includegraphics[width = 1.73cm]{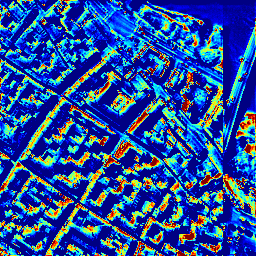}}\!
		\subfloat[{\small Xfeat-TriD}]{\includegraphics[width = 1.73cm]{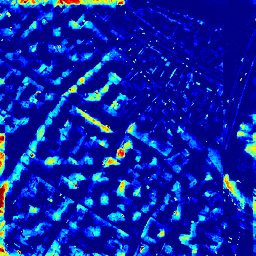}}\!
		\subfloat[{\small Xfeat-ZSL}]{\includegraphics[width = 1.73cm]{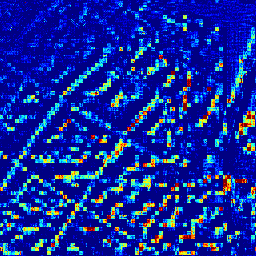}}\!
		\subfloat[{\small SDR-TriD}]{\includegraphics[width = 1.73cm]{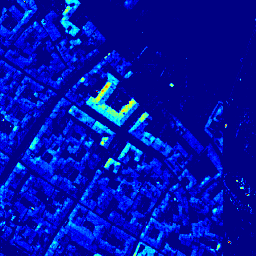}}\!
		\subfloat[{\small SDR-ZSL}]{\includegraphics[width = 1.73cm]{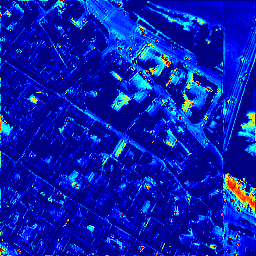}}\!
		\subfloat[{\small SDR-BSF}]{\includegraphics[width = 1.73cm]{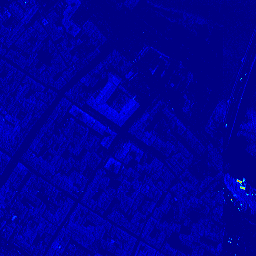}}\!
		\subfloat[{\small GT}]{\includegraphics[width = 1.73cm]{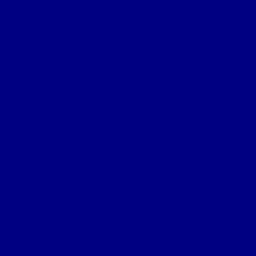}}
		\subfloat{\includegraphics[width = 0.24cm]{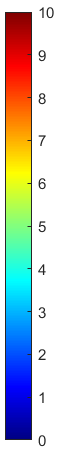}}

		\caption{The fused images and corresponding error images obtained by different methods  for `Pavia Center' (bands 20, 40 and 60,  \textbf{sf} = 4) dataset.\tiny\label{AF}}
	\end{figure*}

	Next, we verify the overall registration and fusion performance of the SDR-BSF method. Specifically, we employ the  blind fusion algorithm TriD \cite{Yang2025} and ZSL \cite{ZSL}  to fuse the pre-registered HSI-MSI  pairs. 
	\begin{table}[hpbt]
		\renewcommand\arraystretch{1.2}
		\caption{The evaluation metrics for the fused images.} 
		\vspace{-0.05cm}      
		\centering
		
		\begin{tabular}{p{0.7cm}p{1.3cm}|p{1.0cm}p{1.0cm}p{1.0cm}p{0.8cm}}
			\hline
			& \textbf{Fusion} & PSNR & SSIM & ERGAS & SAM \\ \hline
			\multicolumn{6}{c}{\textbf{Pavia Center (sf = 4, scaling)}} \\ \hline
			\multirow{2}{*}{SIFT} & TriD \cite{Yang2025}& 31.96 & 0.927 & 5.364 & 7.70 \\ 
			& ZSL \cite{ZSL}& 28.64 & 0.816 & 7.414 & 9.94 \\ \cline{2-6}
			\multirow{2}{*}{NTG} & TriD \cite{Yang2025}& 31.47 & 0.933 & 5.984 & 7.82 \\ 
			& ZSL \cite{ZSL}& 28.98 & 0.829 & 6.977 & 9.67 \\  \cline{2-6}
			\multirow{2}{*}{X-feat} & TriD \cite{Yang2025}& 31.86 & 0.939 & 5.122 & 7.57 \\ 
			& ZSL \cite{ZSL}& 29.63 & 0.843 & 6.737 & 8.91 \\  \cline{2-6}
			\multirow{3}{*}{SDR} & TriD \cite{Yang2025}& 33.80 & 0.952 & 4.389 & 5.11 \\ 
			& ZSL \cite{ZSL}& 32.26 & 0.935 & 5.146 & 7.24 \\ 
			& BSF & \textbf{41.80} & \textbf{0.988} & \textbf{1.773} & \textbf{3.67} \\ 
			\hline
			\multicolumn{6}{c}{\textbf{Superballs (sf = 16, pincushion)}} \\ \hline
			\multirow{2}{*}{SIFT} & TriD \cite{Yang2025}& 26.73 & 0.785 & 6.381 & 32.92 \\ 
			& ZSL \cite{ZSL}& 26.35 & 0.776 & 6.773 & 33.26 \\ \cline{2-6}
			\multirow{2}{*}{NTG} & TriD \cite{Yang2025}& 27.18 & 0.792 & 6.132 & 31.37 \\ 
			& ZSL \cite{ZSL} & 26.46 & 0.779 & 7.196 & 33.85 \\  \cline{2-6}
			\multirow{2}{*}{X-feat} & TriD \cite{Yang2025}& 28.40 & 0.806  & 6.151 & 28.35 \\ 
			& ZSL \cite{ZSL} & 27.78 & 0.785 & 6.834 & 27.74 \\  \cline{2-6}
			\multirow{3}{*}{SDR} & TriD \cite{Yang2025}& 33.96 & 0.925 & 4.238 & 22.74 \\ 
			& ZSL \cite{ZSL} & 32.45 & 0.917 & 5.742 & 25.16 \\   
			& BSF & \textbf{39.78} & \textbf{0.960} & \textbf{1.565} & \textbf{10.09} \\ 
			\hline
		\end{tabular}
		\label{rf}
		\vspace{-2pt}
	\end{table}
	As shown in  Table \ref{rf}, the inadequate registration accuracy of SIFT, NTG, and X-feat results in error accumulation during the subsequent estimation of the degradation operator and fusion stages, ultimately yielding suboptimal performance for these compared approaches. In contrast, the fusion methods achieve superior  performance on the registered images obtained by SDR, which further validates the  registration performance of our approach. Among the compared fusion techniques, TriD outperforms ZSL overall, which may due to TriD simultaneously conducts image registration and fusion, thereby minimizing error accumulation. Notably, the BSF model demonstrates significantly enhanced performance over both TriD and ZSL. This improvement can be attributed to the BSF model's requirement to estimate only the spectral degradation operator \textbf{R}, reducing the ill-posedness of the problem. Next, the fused images and corresponding error maps of the various methods are presented in Figure \ref{AF} for a visual comparison.

	From the figure, the fusion images obtained using SIFT, NTG and X-feat registration exhibit errors primarily in the form of edge stripes and color discrepancies. These issues stem from accumulated errors during registration and degradation operator estimation. In contrast, the fusion images produced with SDR registration show significantly fewer errors. However, the SDR-ZSL method still displays noticeable color differences because the degradation operator estimation in ZSL lacks prior regularization, making it less robust to misregistration. The reconstructed image generated by our SDR-BSF method demonstrates markedly fewer errors, thereby validating the effectiveness of our two-stage approach.

	\subsection{Comparison with other methods}
	
	
	\subsubsection{The experimental results on the simulated data}
	We compare the performance of different mathods under noisy conditions. The SNRh and SNRm are used to denote the signal-to-noise ratio of obtained HSI and MSI, respectively. We set SNRh = 35dB and SNRm = 40dB uniformly.  Table \ref{T0}  displays the experimental results of various methods on the `Pavia'  and the `CAVE' datasets.
	\begin{table*}[hpbt]
		\vspace{0pt}
		\renewcommand\arraystretch{1.1}
		\caption{The experimental results of compared methods on five datasets under different transformations.} 
		\vspace{-0.25cm}      
		\centering
		\label{T0}               
		\begin{threeparttable}
			\begin{tabular}{|p{1.1cm}p{1.73cm}|p{0.8cm}p{0.75cm}p{0.75cm}p{0.75cm}|p{0.8cm}p{0.75cm}p{0.75cm}p{0.75cm}|p{0.8cm}p{0.75cm}p{0.75cm}p{0.75cm}|}
				\bottomrule
				\multicolumn{2}{|c|}{\textbf{Distortion} } & \multicolumn{4}{c|}{\textbf{Scaling} (1.1)} & \multicolumn{4}{c|}{\textbf{Rotation} (2$^\circ$)} & \multicolumn{4}{c|}{\textbf{Pincushion} (1$\times$ 10$^{-2}$)}
				\\ \hline
				\textbf{Dataset} & \textbf{Method} & PSNR & SSIM & ERGAS & SAM & PSNR & SSIM & ERGAS & SAM & PSNR & SSIM & ERGAS & SAM
				\\ \hline
				& Hysure \cite{Hysure} & 35.24 & 0.948 & 3.513 & 5.52 & 36.75 & 0.961 & 2.632 & 4.76 & 36.04 & 0.954 & 3.301 & 5.73 \\
				& Integrated \cite{Zhou2020}& 31.85 & 0.904 & 5.558 & 6.36 & 32.40 & 0.916 & 5.104 & 6.19 & 31.68 & 0.897 & 5.359 & 6.82 \\
				& NED \cite{NED}& 35.45 & 0.957 & 3.072 & 4.38 & 34.40 & 0.949 & 3.386 & 4.54 & 33.77 & 0.942 & 3.776 & 5.15 \\
				\textbf{Pavia C}& NonregSR  \cite{Zheng2022}& 34.09 & 0.940 & 3.752 & 5.49 & 35.68 & 0.952 & 3.316 & 4.98 & 34.66 & 0.945 & 3.634 & 5.23 \\
				(\textbf{sf} = 4)& PMIRFCo \cite{PMI}& 37.16 & 0.962 & 2.936 & 4.96 & 37.57 & 0.964 & 2.611 & 4.81 & 37.65 & 0.965 & 2.627 & 4.73 \\
				& DFMF \cite{Guo2022}& 38.56 & 0.974 & 2.234 & 4.14 & 38.11 & 0.972 & 2.274 & 4.26 & 38.48 & 0.974 & 2.226 & 4.21 \\
				& HPWRL  \cite{Nie2024}& 37.62 & 0.963 & 2.551 & 4.63 & 37.80 & 0.963 & 2.578 & 4.59 & 37.44 & 0.961 & 2.775 & 4.68 \\
				& IR-ArF \cite{Qu2025}& 39.14 & 0.976 & 2.175 & 4.27 & 38.31 & 0.972 & 2.159 & 4.32 & 38.06 & 0.970 & 2.137 & 4.28 \\
				& SDR-BSF & \textbf{41.58} & \textbf{0.985} & \textbf{1.842} & \textbf{3.87} & \textbf{40.74} & \textbf{0.985} & \textbf{1.711} & \textbf{3.68} & \textbf{40.25} & \textbf{0.978} & \textbf{1.916} & \textbf{3.75} \\
				\cline{1-14}
				& Hysure \cite{Hysure} & 34.16 & 0.946 & 1.447 & 3.63 & 34.93 & 0.958 & 1.282 & 3.37 & 35.68 & 0.960 & 1.116 & 3.09 \\
				& Integrated \cite{Zhou2020}& 30.24 & 0.868 & 3.382 & 5.35 & 31.33 & 0.879 & 3.159 & 5.16 & 31.76 & 0.875 & 3.574 & 4.84 \\
				& NED  \cite{NED}& 35.81 & 0.937 & 1.291 & 4.41 & 36.61 & 0.943 & 1.364 & 4.70 & 36.13 & 0.942 & 1.236 & 4.15 \\
				\textbf{Pavia U}& NonregSR  \cite{Zheng2022}& 32.91 & 0.932 & 1.603 & 3.28 & 32.33 & 0.935 & 1.739 & 3.25 & 32.82 & 0.943 & 1.655 & 3.12 \\
				(\textbf{sf} = 8)& PMIRFCo  \cite{PMI}& 36.34 & 0.960 & 1.143 & 2.94 & 36.68 & 0.961 & 1.042 & 2.91 & 37.01 & 0.962 & 1.023 & 2.84 \\
				& DFMF \cite{Guo2022}& 36.92 & 0.965 & 1.072 & 3.55 & 37.21 & 0.969 & 0.973 & 3.46 & 36.40 & 0.959 & 1.150 & 3.73 \\
				& HPWRL  \cite{Nie2024}& 35.47 & 0.962 & 1.470 & 3.71 & 35.16 & 0.966 & 1.322 & 3.72& 35.99 & 0.968 & 1.310 & 3.64 \\
				& IR-ArF \cite{Qu2025}& 37.25 & 0.969 & 1.075 & 2.93 & 37.11 & 0.968 & 1.083 & 2.96 & 37.78 & 0.970 & 1.021 & 2.86 \\
				& SDR-BSF & \textbf{40.08} & \textbf{0.983} & \textbf{0.803} & \textbf{2.66} & \textbf{39.54} & \textbf{0.981} & \textbf{0.854} & \textbf{2.82} & \textbf{40.12} & \textbf{0.984} & \textbf{0.769} & \textbf{2.56} \\
				\hline
				& Hysure \cite{Hysure} & 34.65 & 0.893 & 2.957 & 17.36 & 35.01 & 0.906 & 2.826 & 16.90 & 34.88 & 0.901 & 2.916 & 16.79 \\
				& Integrated \cite{Zhou2020}& 32.64 & 0.901 & 3.762 & 19.12 & 32.23 & 0.896 & 3.559 & 20.66 & 31.96 & 0.905 & 5.324 & 20.93 \\
				& NED  \cite{NED}& 36.78 & 0.959 & 2.032 & 16.85 & 36.97 & 0.960 & 1.983 & 16.59 & 36.39 & 0.956 & 2.557 & 15.41 \\
				~\textbf{Face}& NonregSR  \cite{Zheng2022}& 34.61 & 0.948 & 3.352 & 16.52 & 35.25 & 0.884 & 2.471 & 15.76 & 34.47 & 0.944 & 3.490 & 17.14 \\
				(\textbf{sf} = 8)& PMIRFCo  \cite{PMI}& 37.47 & 0.960 & 2.436 & 13.84 & 37.96 & 0.961 & 2.267 & 13.86 & 36.82 & 0.961 & 2.786 & 14.32 \\
				& DFMF \cite{Guo2022}& 37.32 & 0.947 & 2.195 & 15.83 & 37.68 & 0.952 & 2.101 & 15.15 & 37.44 & 0.951 & 2.157 & 15.29 \\
				& HPWRL  \cite{Nie2024}& 38.21 & 0.959 & 1.922 & 14.16 & 38.07 & 0.961 & 1.944 & 14.25 & 37.89 & 0.958 & 2.275 & 14.66 \\
				& IR-ArF \cite{Qu2025}& 39.43 & 0.971 & 1.879 & 13.75 & 39.36 & 0.971 & 1.884 & 13.94 & 38.77 & 0.968 & 2.063 & 14.15 \\
				& SDR-BSF & \textbf{41.28} & \textbf{0.975} & \textbf{1.625} & \textbf{12.51} & \textbf{41.56} & \textbf{0.976} & \textbf{1.610} & \textbf{11.86} & \textbf{40.82} & \textbf{0.969} & \textbf{1.818} & \textbf{12.44} \\
				\cline{1-14}
				& Hysure \cite{Hysure} & 26.87 & 0.718 & 5.041 & 36.92 & 27.34 & 0.732 & 4.742 & 32.25 & 26.59 & 0.724 & 4.887 & 34.93 \\
				& Integrated \cite{Zhou2020}& 28.17 & 0.743 & 4.739 & 34.98 & 28.35 & 0.752 & 4.652 & 31.92 & 27.78 & 0.748 & 4.663 & 32.29 \\
				& NED  \cite{NED}& 32.96 & 0.815 & 3.259 & 26.67 & 32.59 & 0.806 & 3.468 & 27.57 & 32.93 & 0.810 & 3.351 & 27.44 \\
				\textbf{Superballs}& NonregSR  \cite{Zheng2022}& 32.41 & 0.822 & 4.205 & 29.88 & 32.17 & 0.818 & 4.437 & 30.14 & 32.01 & 0.817 & 4.254 & 30.89 \\
				(\textbf{sf} = 16)&PMIRFCo \cite{PMI}& 36.22 & 0.924 & 2.073 & 15.14 & 36.39 & 0.925 & 1.989 & 15.03 & 35.95 & 0.924 & 2.109 & 15.40  \\
				& DFMF \cite{Guo2022}& 35.64 & 0.912 & 2.415 & 16.85 & 36.47 & 0.923 & 2.389 & 15.60 & 36.15 & 0.920 & 2.426 & 16.28 \\
				& HPWRL  \cite{Nie2024}& 35.21 & 0.918 & 2.689 & 15.58 & 36.14 & 0.935 & 2.301 & 14.89 & 35.73 & 0.927 & 2.570 & 15.32 \\
				& IR-ArF \cite{Qu2025}& 37.57 & \textbf{0.950} & 1.937 & 13.62 & 37.55 & 0.950 & 1.954 & 13.89 & 37.31 & 0.949 & 2.074 & 14.56 \\
				& SDR-BSF & \textbf{39.13} & 0.943 & \textbf{1.726} & \textbf{11.89} & \textbf{40.05} & \textbf{0.969} & \textbf{1.496} & \textbf{10.31} & \textbf{39.65} & \textbf{0.952} & \textbf{1.658} & \textbf{10.67} \\
				\cline{1-14}
				& Hysure \cite{Hysure} & 24.90 & 0.791 & 1.217 & 34.20 & 25.54 & 0.768 & 1.281 & 35.68 & 26.28 & 0.812 & 1.121 & 32.16 \\
				& Integrated \cite{Zhou2020}& 27.72 & 0.906 & 1.169 & 22.14 & 28.91 & 0.909 & 1.086 & 22.07 & 29.23 & 0.905 & 1.067 & 22.40 \\
				& NED  \cite{NED}& 33.54 & 0.937 & 1.062 & 17.05 & 34.00 & 0.945 & 0.960 & 16.74 & 34.68 & 0.952 & 0.974 & 16.20 \\
				~\textbf{Toys}& NonregSR  \cite{Zheng2022}& 30.76 & 0.801 & 1.196 & 23.35 & 29.93 & 0.817 & 1.294 & 21.87 & 30.47 & 0.822 & 1.173 & 20.99 \\
				(\textbf{sf} = 32)&  PMIRFCo  \cite{PMI}& 34.65 & 0.941 & 0.897 & 16.43 & 35.42 & 0.946 & 0.972 & 21.71 & 34.27 & 0.911 & 3.471 & 15.74\\
				& DFMF \cite{Guo2022}& 35.67 & 0.947 & 0.934 & 14.74 & 35.83 & 0.947 & 0.879 & 15.27 & 36.08 & 0.948 & 0.933 & 15.49 \\
				& HPWRL  \cite{Nie2024}& 33.68 & 0.943 & 1.068 & 16.96 & 34.16 & 0.945 & 1.045 & 16.59 & 33.35 & 0.942 &1.027  & 17.11 \\
				& IR-ArF \cite{Qu2025}& 35.02 & 0.945 & 0.916 & 15.37 & 34.95 & 0.944 & 0.923 & 15.54 & 35.27 & 0.947 & 0.881 & 15.07 \\
				& SDR-BSF & \textbf{37.85} & \textbf{0.954} & \textbf{0.699} & \textbf{13.05} & \textbf{38.03} & \textbf{0.954} & \textbf{0.733} & \textbf{13.21} & \textbf{38.19} & \textbf{0.954} & \textbf{0.658} & \textbf{12.96} \\
				\bottomrule
			\end{tabular}
		\end{threeparttable}
		\vspace{-0pt}
	\end{table*}

\begin{figure*}[htbp]
		\captionsetup[subfloat]{labelsep=none,format=plain,labelformat=empty}
		\centering
		\vspace{-10pt}
		\subfloat{\includegraphics[width=1.734cm]{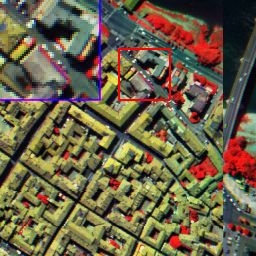}} \!
		\subfloat{\includegraphics[width=1.734cm]{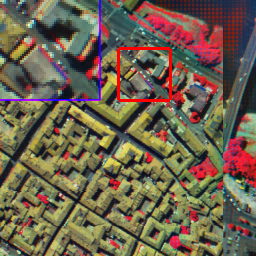}} \!
		\subfloat{\includegraphics[width=1.734cm]{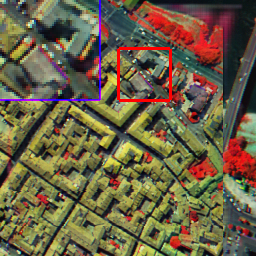}} \!
		\subfloat{\includegraphics[width=1.734cm]{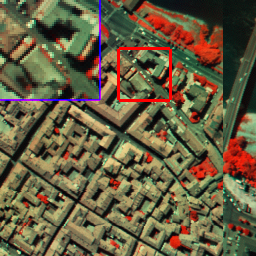}} \!
		\subfloat{\includegraphics[width=1.734cm]{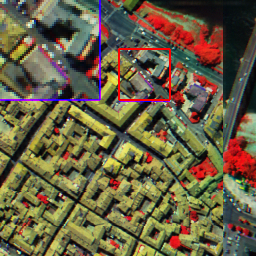}} \!
		\subfloat{\includegraphics[width=1.734cm]{SSDR/SPL_BRF_PC.png}}\!
		\subfloat{\includegraphics[width=1.734cm]{SSDR/GT_PC.png}}\!
		\subfloat{\includegraphics[width=1.734cm]{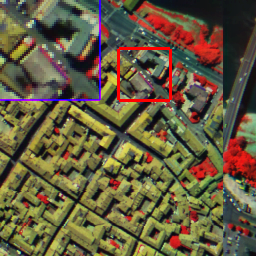}}\!
		\subfloat{\includegraphics[width=1.734cm]{SSDR/SPL_BRF_PC.png}}\!
		\subfloat{\includegraphics[width=1.734cm]{SSDR/GT_PC.png}}
		\subfloat{\includegraphics[width = 0.24cm]{SSDR/k.png}}
		
		\vspace{-9pt}
		\subfloat{\includegraphics[width=1.734cm]{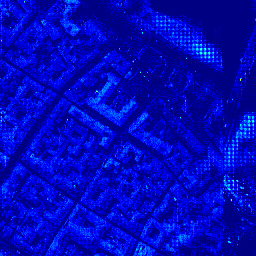}}\!
		\subfloat{\includegraphics[width=1.734cm]{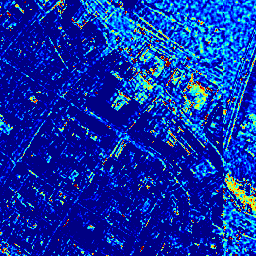}}\!
		\subfloat{\includegraphics[width=1.734cm]{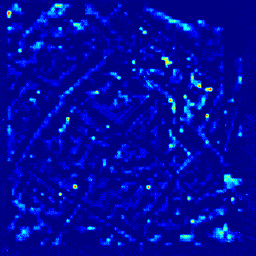}}\!
		\subfloat{\includegraphics[width=1.734cm]{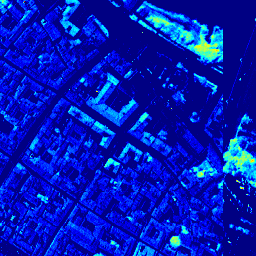}}\!
		\subfloat{\includegraphics[width=1.734cm]{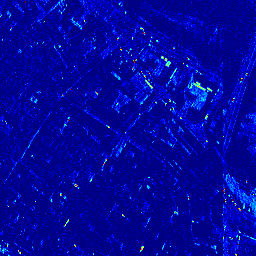}}\!
		\subfloat{\includegraphics[width=1.734cm]{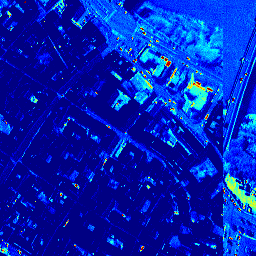}}\!
		\subfloat{\includegraphics[width=1.734cm]{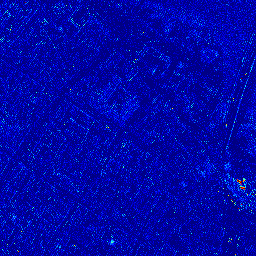}}\!
		\subfloat{\includegraphics[width=1.734cm]{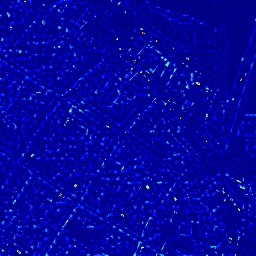}}\!
		\subfloat{\includegraphics[width=1.734cm]{SSDR/SPL_BRF_PCT.png}}\!
		\subfloat{\includegraphics[width=1.734cm]{SSDR/PaviaT_GT.png}}
		\subfloat{\includegraphics[width = 0.24cm]{SSDR/bband.png}}
		\vspace{-8pt}
		
		\subfloat{\includegraphics[width=1.734cm]{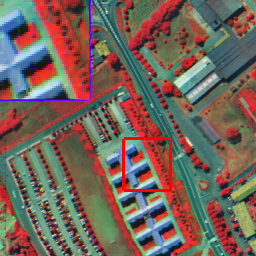}}\!
		\subfloat{\includegraphics[width=1.734cm]{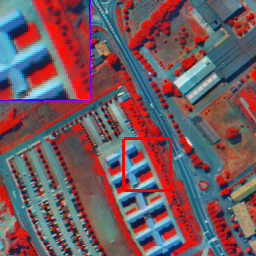}} \!
		\subfloat{\includegraphics[width=1.734cm]{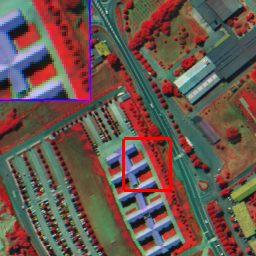}}\!
		\subfloat{\includegraphics[width=1.734cm]{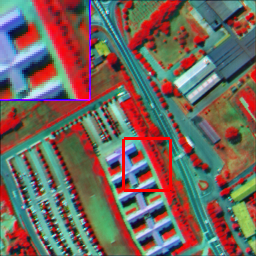}} \!
		\subfloat{\includegraphics[width=1.734cm]{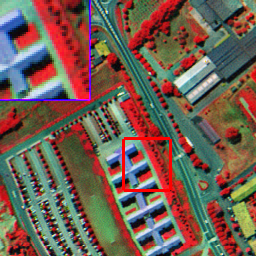}} \!
		\subfloat{\includegraphics[width=1.734cm]{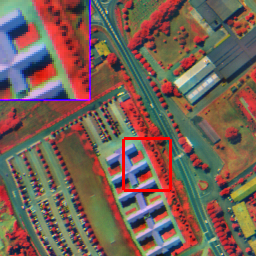}} \!
		\subfloat{\includegraphics[width=1.734cm]{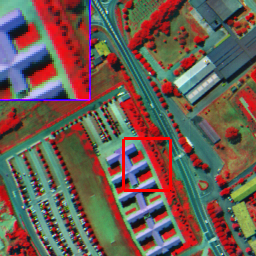}} \!
		\subfloat{\includegraphics[width=1.734cm]{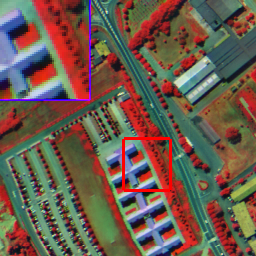}}\!
		\subfloat{\includegraphics[width=1.734cm]{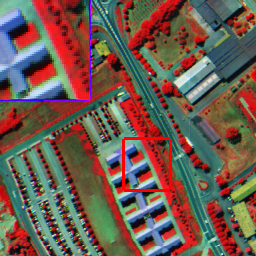}}\!
		\subfloat{\includegraphics[width=1.734cm]{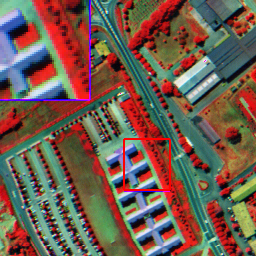}}
		\subfloat{\includegraphics[width = 0.24cm]{SSDR/k.png}}
		
		\vspace{-9pt}
		\subfloat[{\small Hysure}]{\includegraphics[width=1.734cm]{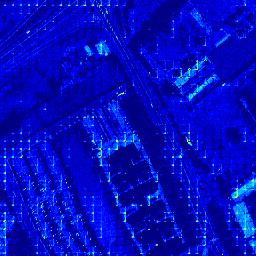}}\!
		\subfloat[{\small Integrated}]{\includegraphics[width=1.734cm]{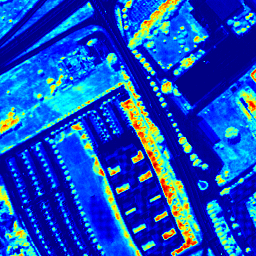}}\!
		\subfloat[{\small NED}]{\includegraphics[width=1.734cm]{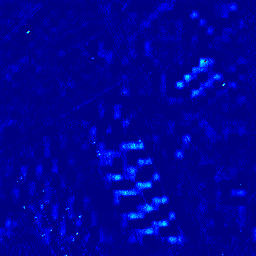}}\!
		\subfloat[{\small NonregSR}]{\includegraphics[width=1.734cm]{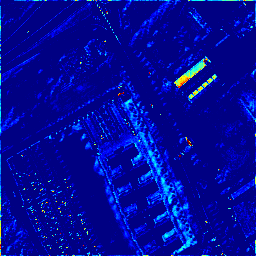}}\!
		\subfloat[{\small PMIRFCo}]{\includegraphics[width=1.734cm]{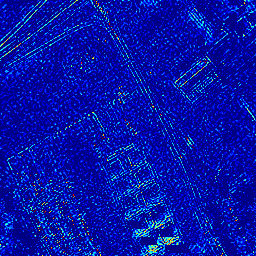}}\!
		\subfloat[{\small DFMF}]{\includegraphics[width=1.734cm]{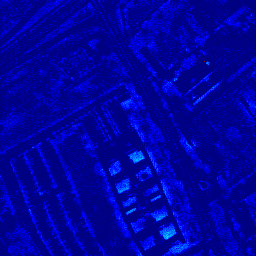}}\!
		\subfloat[{\small HPWRL}]{\includegraphics[width=1.734cm]{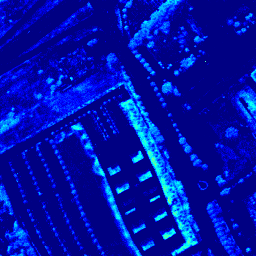}}\!
		\subfloat[{\small IR-ArF}]{\includegraphics[width=1.734cm]{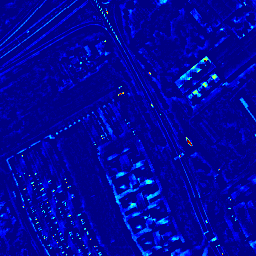}}\!
		\subfloat[{\small SDR-BSF}]{\includegraphics[width=1.734cm]{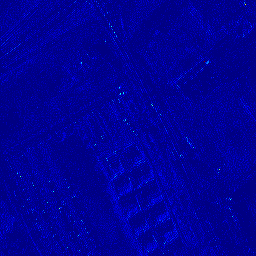}}\!
		\subfloat[{\small GT}]{\includegraphics[width=1.734cm]{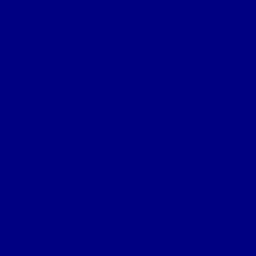}}
		\subfloat{\includegraphics[width = 0.24cm]{SSDR/bband.png}}
		
		\vspace{-7pt}
		\subfloat{\includegraphics[width=1.734cm]{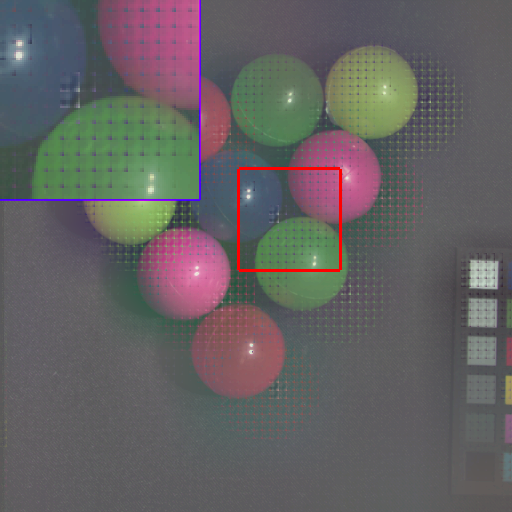}}\!
		\subfloat{\includegraphics[width=1.734cm]{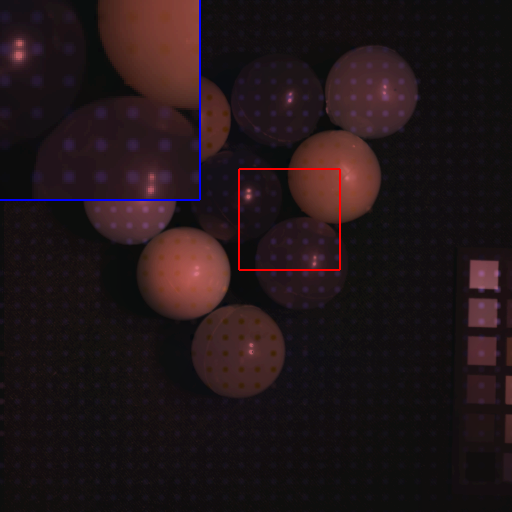}} \!
		\subfloat{\includegraphics[width=1.734cm]{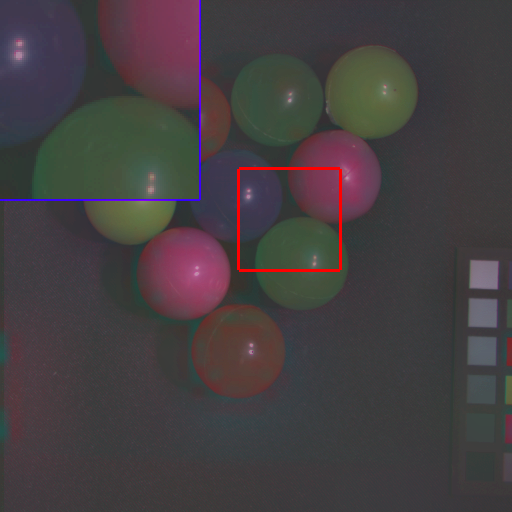}}\!
		\subfloat{\includegraphics[width=1.734cm]{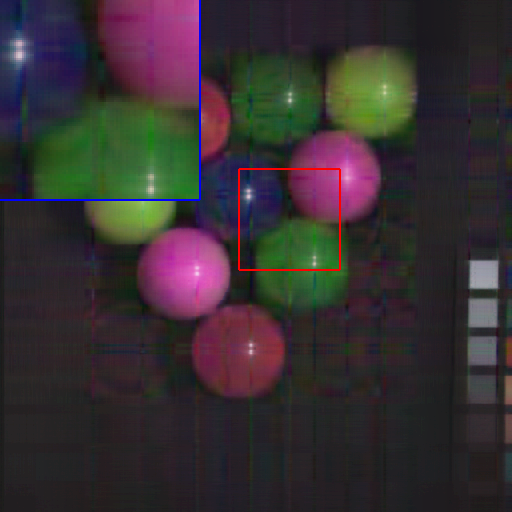}} \!
		\subfloat{\includegraphics[width=1.734cm]{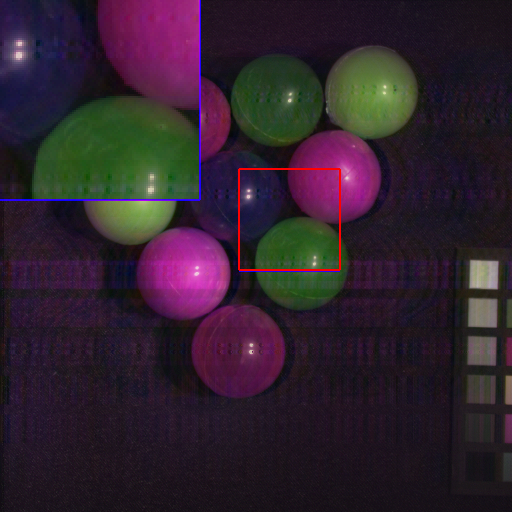}} \!
		\subfloat{\includegraphics[width=1.734cm]{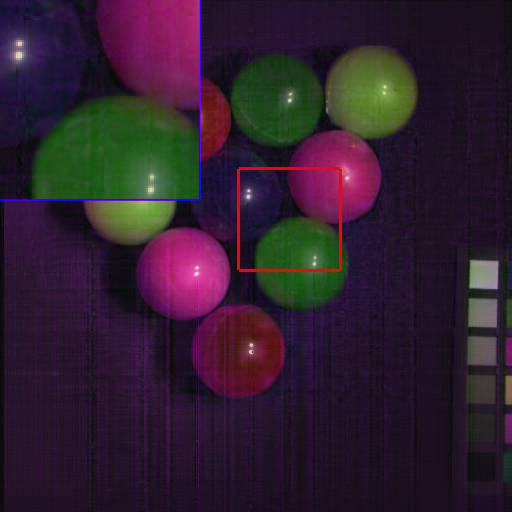}} \!
		\subfloat{\includegraphics[width=1.734cm]{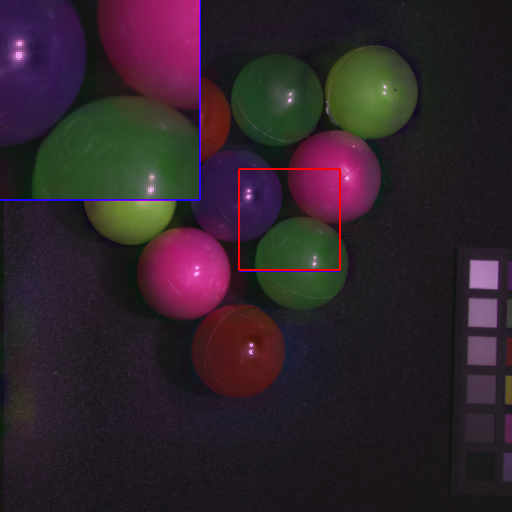}} \!
		\subfloat{\includegraphics[width=1.734cm]{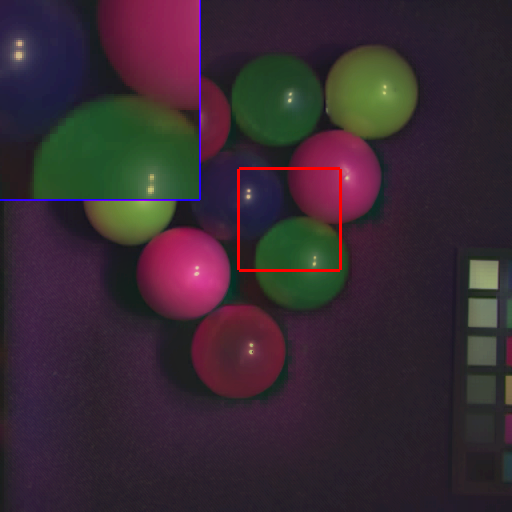}}\!
		\subfloat{\includegraphics[width=1.734cm]{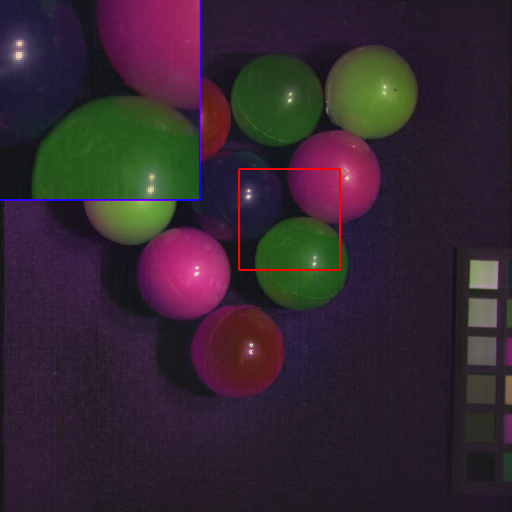}}\!
		\subfloat{\includegraphics[width=1.734cm]{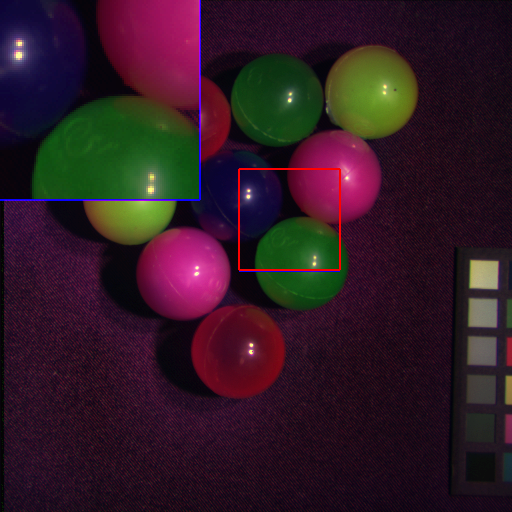}}
		\subfloat{\includegraphics[width = 0.24cm]{SSDR/k.png}}
		
		\vspace{-9pt}
		\subfloat{\includegraphics[width=1.734cm]{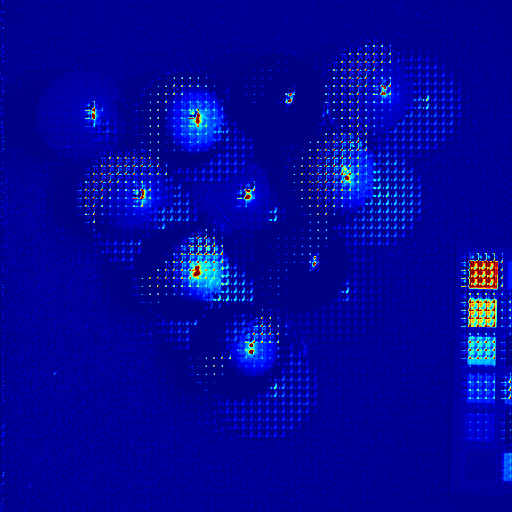}}\!
		\subfloat{\includegraphics[width=1.734cm]{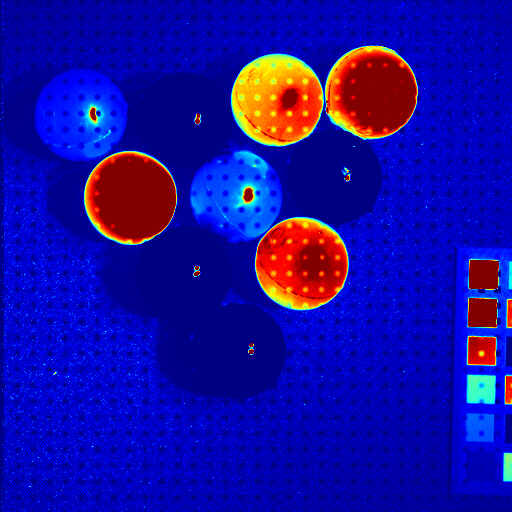}}\!
		\subfloat{\includegraphics[width=1.734cm]{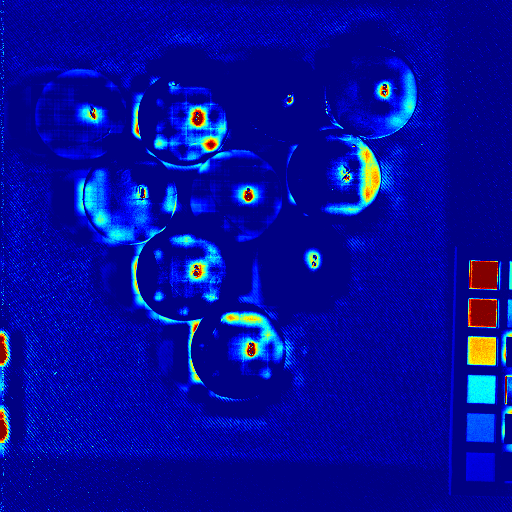}}\!
		\subfloat{\includegraphics[width=1.734cm]{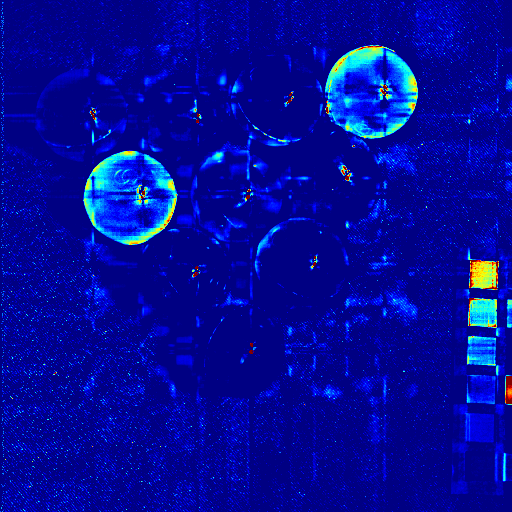}}\!
		\subfloat{\includegraphics[width=1.734cm]{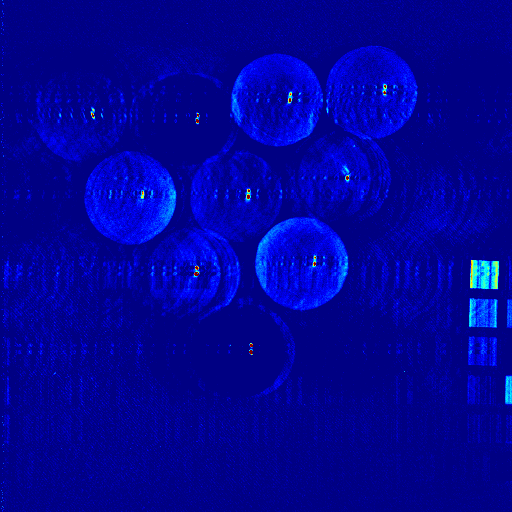}}\!
		\subfloat{\includegraphics[width=1.734cm]{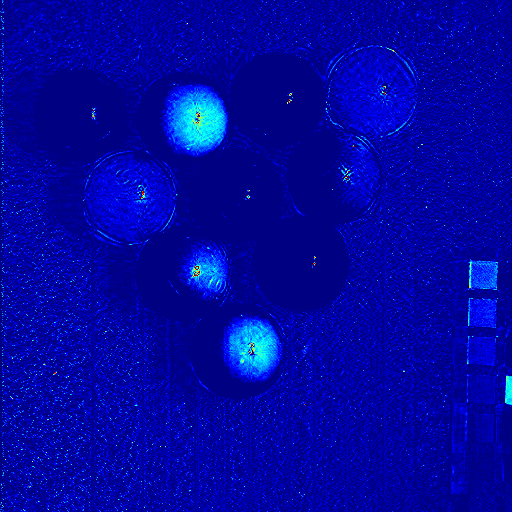}}\!
		\subfloat{\includegraphics[width=1.734cm]{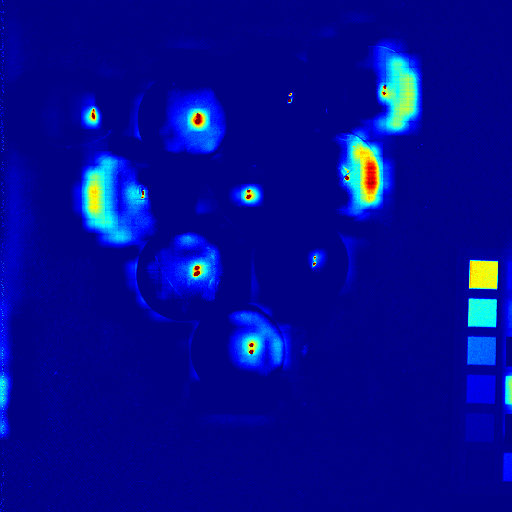}}\!
		\subfloat{\includegraphics[width=1.734cm]{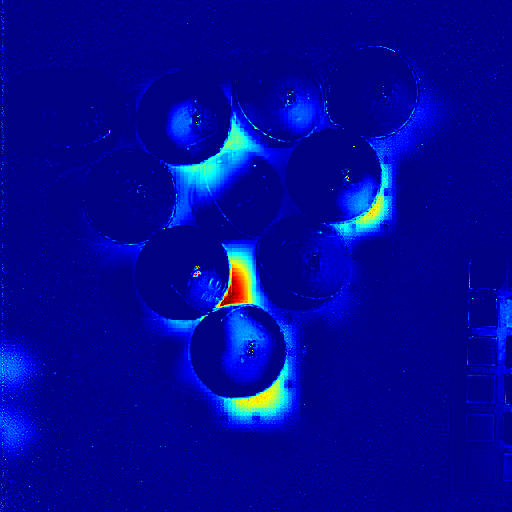}}\!
		\subfloat{\includegraphics[width=1.734cm]{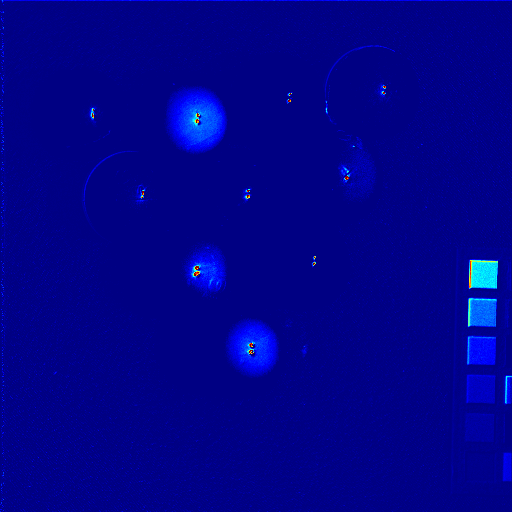}}\!
		\subfloat{\includegraphics[width=1.734cm]{SSDR/PU_GTT.png}}
		\subfloat{\includegraphics[width = 0.24cm]{SSDR/bband.png}}
		
		\vspace{-7pt}
		
		\subfloat{\includegraphics[width=1.734cm]{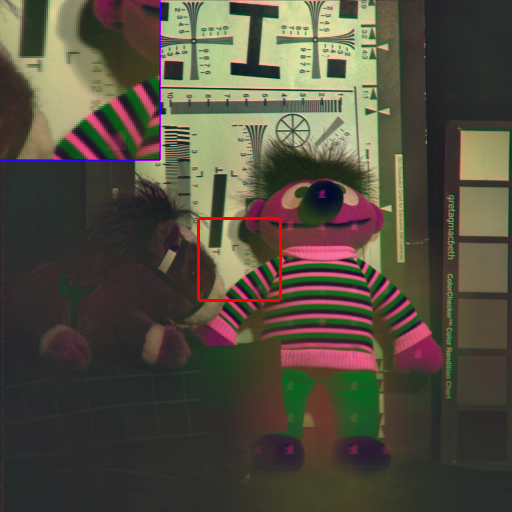}} \!
		\subfloat{\includegraphics[width=1.734cm]{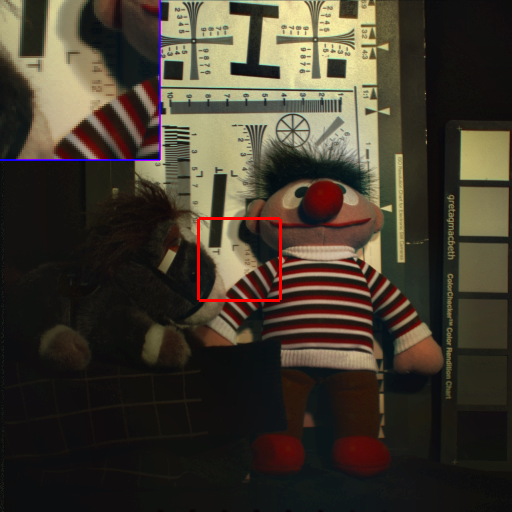}} \!
		\subfloat{\includegraphics[width=1.734cm]{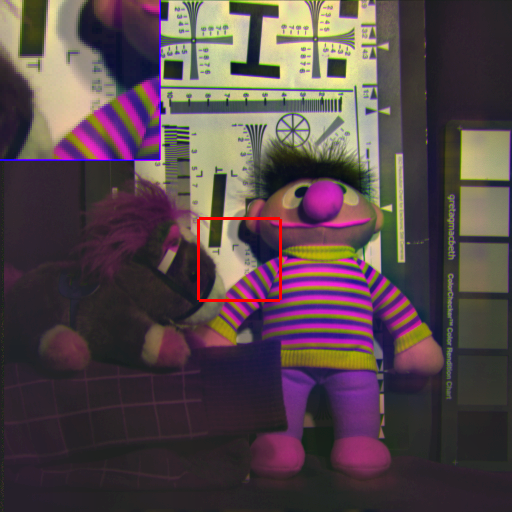}} \!
		\subfloat{\includegraphics[width=1.734cm]{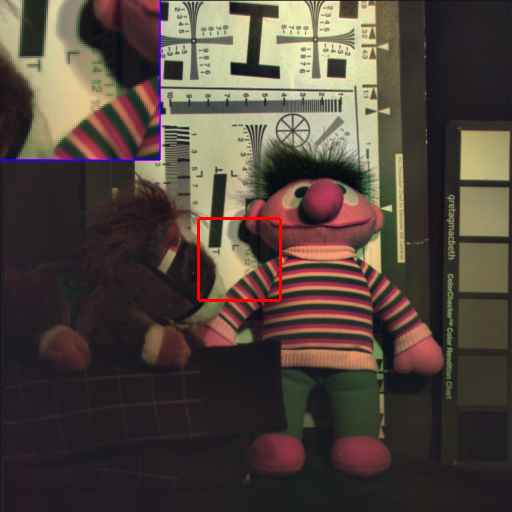}} \!
		\subfloat{\includegraphics[width=1.734cm]{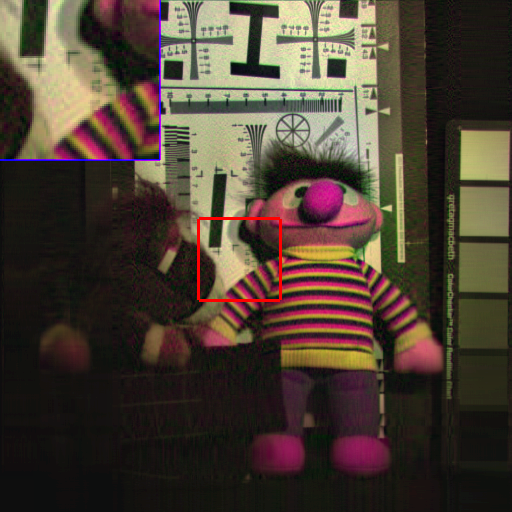}} \!
		\subfloat{\includegraphics[width=1.734cm]{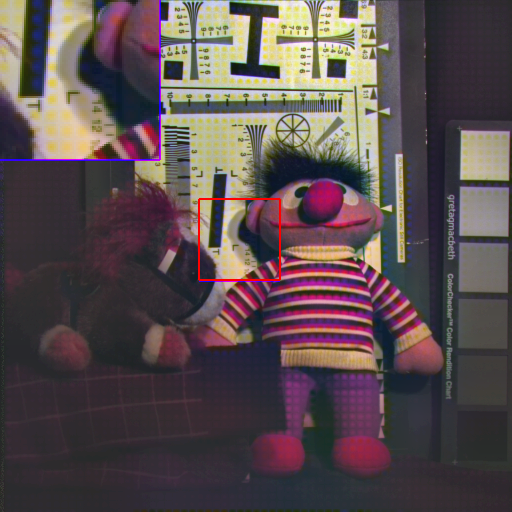}}\!
		\subfloat{\includegraphics[width=1.734cm]{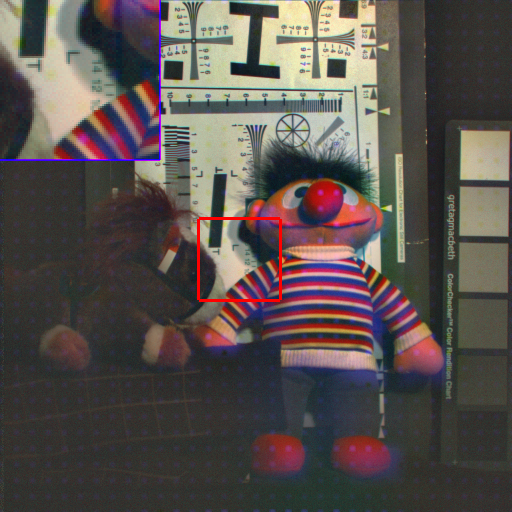}}\!
		\subfloat{\includegraphics[width=1.734cm]{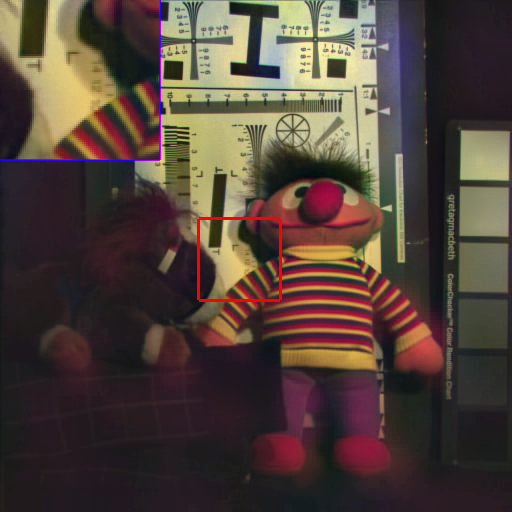}}\!
		\subfloat{\includegraphics[width=1.734cm]{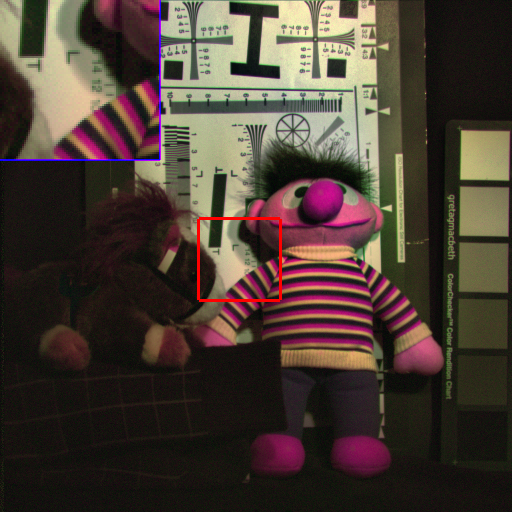}}\!
		\subfloat{\includegraphics[width=1.734cm]{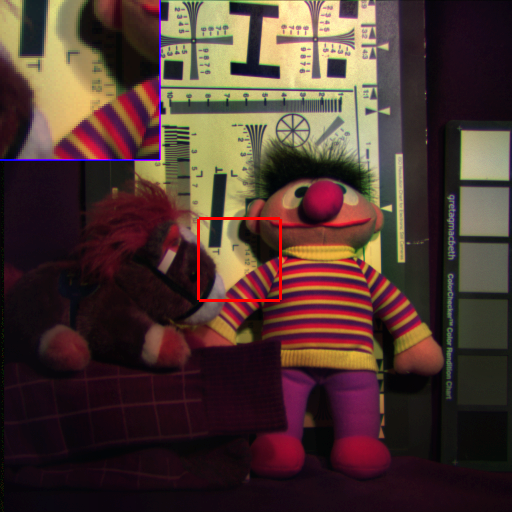}}
		\subfloat{\includegraphics[width = 0.24cm]{SSDR/k.png}}
		
		\vspace{-9pt}
		\subfloat[{\small Hysure}]{\includegraphics[width=1.734cm]{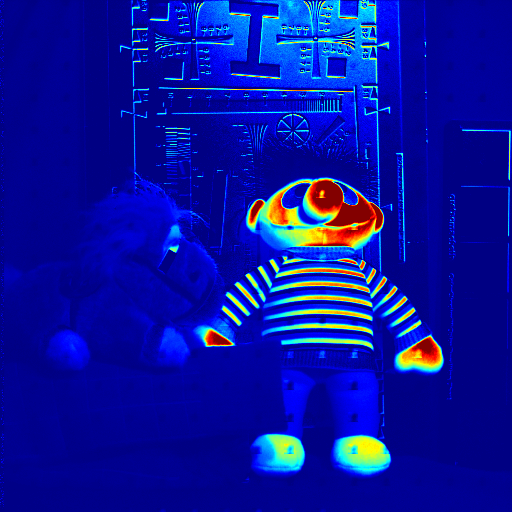}}\!
		\subfloat[{\small Integrated}]{\includegraphics[width=1.734cm]{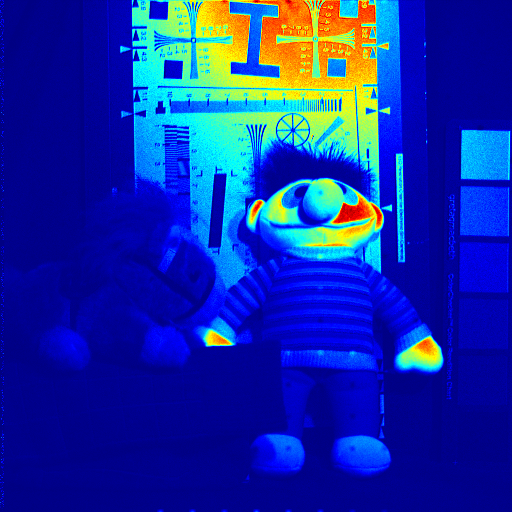}}\!
		\subfloat[{\small NED}]{\includegraphics[width=1.734cm]{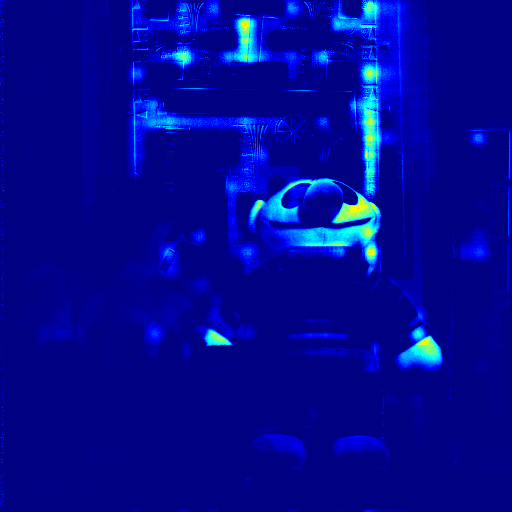}}\!
		\subfloat[{\small NonregSR}]{\includegraphics[width=1.734cm]{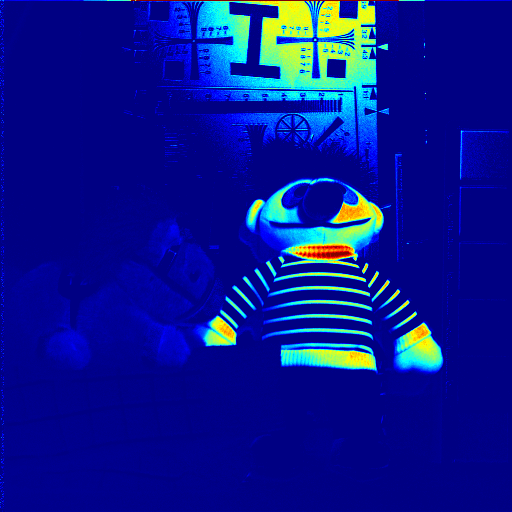}}\!
		\subfloat[{\small PMIRFCo}]{\includegraphics[width=1.734cm]{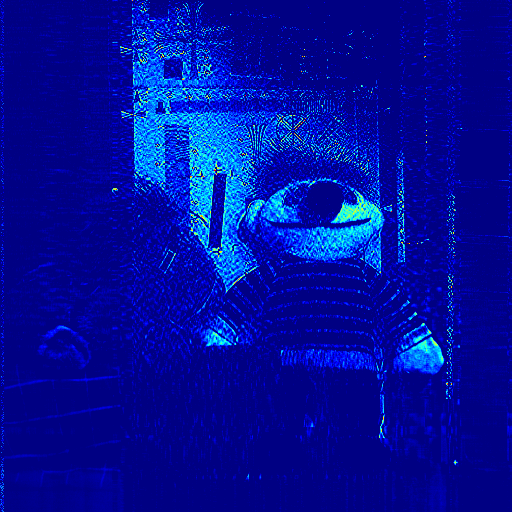}}\!
		\subfloat[{\small DFMF}]{\includegraphics[width=1.734cm]{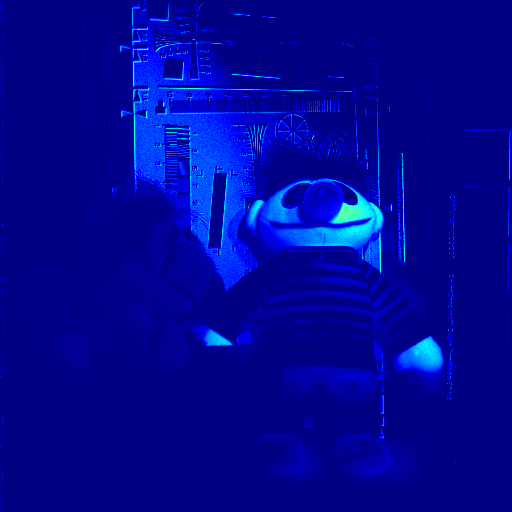}}\!
		\subfloat[{\small HPWRL}]{\includegraphics[width=1.734cm]{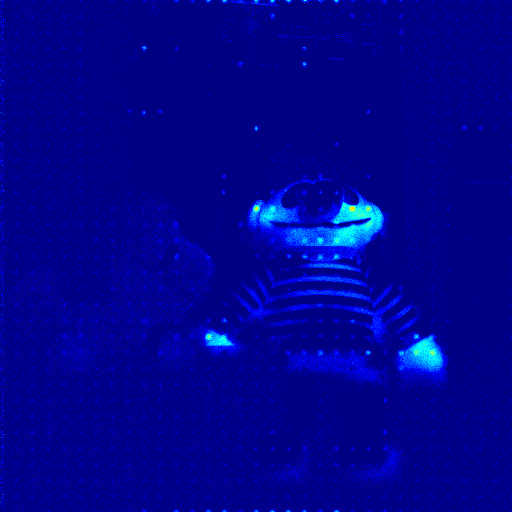}}\!
		\subfloat[{\small IR-ArF}]{\includegraphics[width=1.734cm]{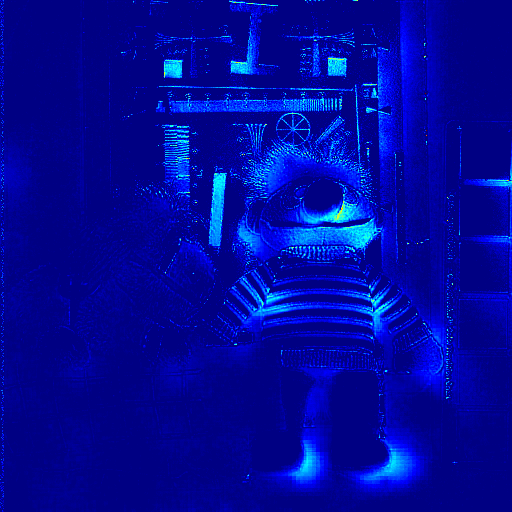}}\!
		\subfloat[{\small SDR-BSF}]{\includegraphics[width=1.734cm]{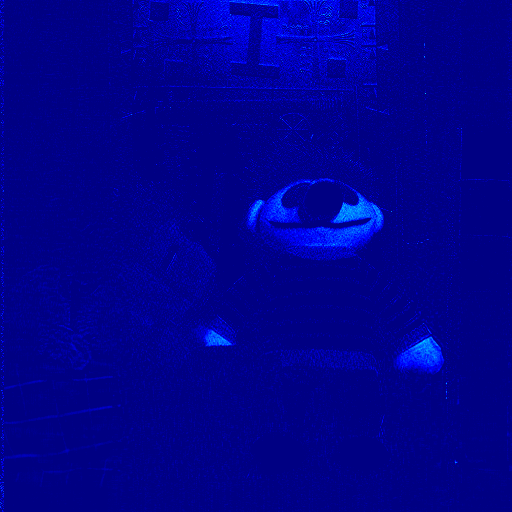}}\!
		\subfloat[{\small GT}]{\includegraphics[width=1.734cm]{SSDR/PaviaT_GT.png}}
		\subfloat{\includegraphics[width = 0.24cm]{SSDR/bband.png}}
		
		
		\caption{Visualization of the fused images  and error maps  for  the  `Pavia' (bands 20, 40 and 60) and   `CAVE' (bands 10, 20, and 30) datasets. The `Pavia Center' and `Superballs' images undergoe scaling distortion, whereas the `Pavia University' and `Toys' dataset undergoe pincushion transformation.}
		\label{DF11}
		\vspace{-2pt}
	\end{figure*}

    From the tables,  `Hysure'  exhibits a certain level of robustness to misregistration despite not including a registration process. Its performance remains competitive when the scale factor is small. This suggests that Total Variation regularization can mitigate the impact of misalignment to some extent. The `Integrated' method does not show ideal performance on all the  datasets. A possible reason is that `Integrated' relies entirely on optimization modeling. In scenarios where images are unregistered and degradation operators are unknown, there is insufficient known information and too many parameters to estimate. The vast solution space makes it challenging for the algorithm to find an optimal solution. The `NED'  adopts a method of registration followed by fusion, which may lead to  accumulation of errors. Therefore, its reconstruction effect generally cannot achieve the optimal result. In summary, owing to insufficient information, model-driven methods tend to exhibit significant ill-posedness, leading to suboptimal performance. In contrast, deep learning-based methods generally achieve superior fusion performance, largely due to their powerful feature extraction capabilities and the ability to integrate registration and fusion modules synergistically. The `PMIRFCo' method leverages multi-scale information within images, performing registration and fusion across multiple scales to achieve more precise alignment and enhanced fusion quality. The `NonregSR' exhibits a slight performance advantage over the `Integrated' approach. However, its effectiveness is unsatisfactory at higher scale factor, such as in the CAVE dataset.   `DFMF' relies on geolocation information for pre-registration, and for non-geographical HSIs, its fusion performance also demonstrates a competitive performance. `HPWRL'  achieved better reconstruction performance when the scale factor is small.  `IR-ArF'  overall demonstrates strong competitiveness. It translates each step of the optimization algorithm into deep network modules and jointly incorporates implicit neural networks to flexibly adjust the image resolution. However, it still lacks  robustness in the case of large scale factors.
	In comparison, our method consistently achieves optimal restoration performance, which is mainly owing to the innovative strategy of using spectral super-resolution on the MSI to obtain registered images. Local misregistration and significant downsampling primarily degrade the spatial information of HSIs, with considerably less impact on spectral information. Consequently, our spectral reconstruction-based approach demonstrates exceptional robustness. Additionally, the BSF method effectively leverages prior information from the images, delivering excellent semi-blind fusion performance while reducing computational costs. Next,  we display the reconstructed images and corresponding error maps for each method in Figure \ref{DF11}.  
	
	As shown in the figures, `Hysure' exhibits noticeable color differences and stripes on the `CAVE' dataset, although its performance on the `Pavia' dataset is relatively superior. The reconstructed images from `Integrated' and `NED' also show significant distortions and color discrepancies in the `CAVE' dataset.  These issues primarily stem from inaccuracies in the estimation of the degradation operators \textbf{B} and \textbf{R}. In summary, model-driven methods tend to perform less effectively at larger scale factors. `NonregSR' performs well on the `Pavia' dataset but exhibits noticeable flaws on the `CAVE'. This may be due to a significant decrease in registration performance when the scale factor is large, which subsequently leads to suboptimal fusion performance. In comparison, `PMIRFCo' exhibits significantly fewer flaws, which can be attributed to its multi-scale learning approach. This method fully exploits image details and enhances the network's learning accuracy, resulting in superior performance. The reconstruction images from `DFMF' also show slight distortions, likely due to less precise estimation of degradation operators when registration and degradation are estimated simultaneously. The fused images of the `Superballs'  using the `HPWRL' method display localized distortions.  From the error map, it can be seen that the defects of `IR-ArF' are mainly concentrated in the edge regions of the image. This also indicates that the registration performance of `IR-ArF' is not robust enough under large spatial scle factors.
	In contrast, the SDR-BSF method leverages spectral super-resolution, ensuring high-quality fusion as long as the spectral information is complete, which allows  to maintain robust performance even at larger scale factors. Consequently, our method produces fused images with higher quality and notably fewer artifacts in the error maps compared to other methods, which  verifies the effectiveness of our approach.

	\subsubsection{The experimental results on real-world data}
\begin{figure*}[htbp]
		\captionsetup[subfloat]{labelsep=none,format=plain,labelformat=empty}
		\centering
		\vspace{-12pt}
		\subfloat[HSI]{\includegraphics[width=3.5cm]{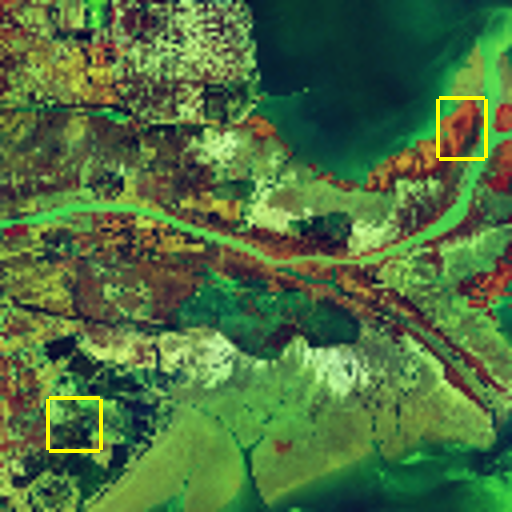}}\!
		\subfloat[MSI]{\includegraphics[width=3.5cm]{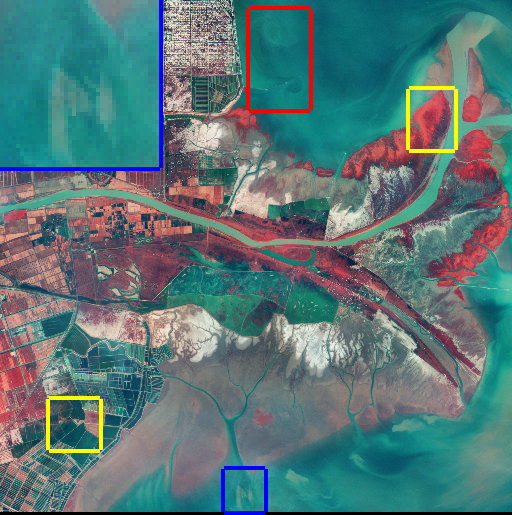}}\!
		\subfloat[Integrated]{\includegraphics[width=3.5cm]{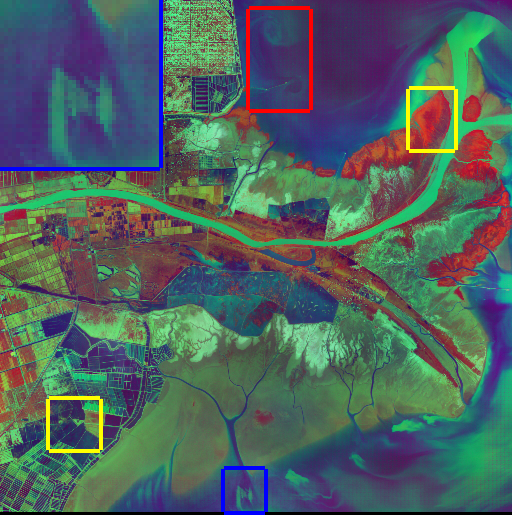}}\!
		\subfloat[NED]{\includegraphics[width=3.5cm]{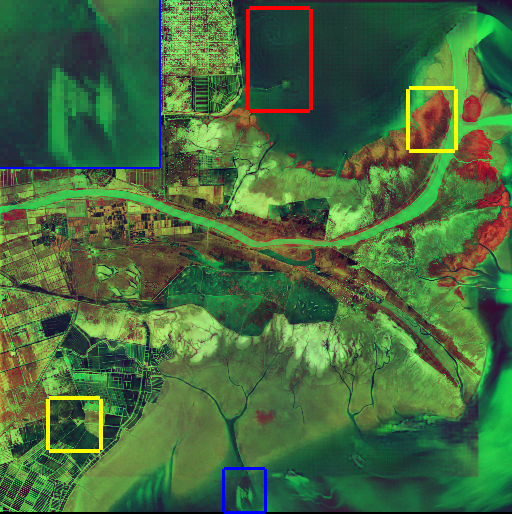}}\!
		\subfloat[NonregSR]{\includegraphics[width=3.5cm]{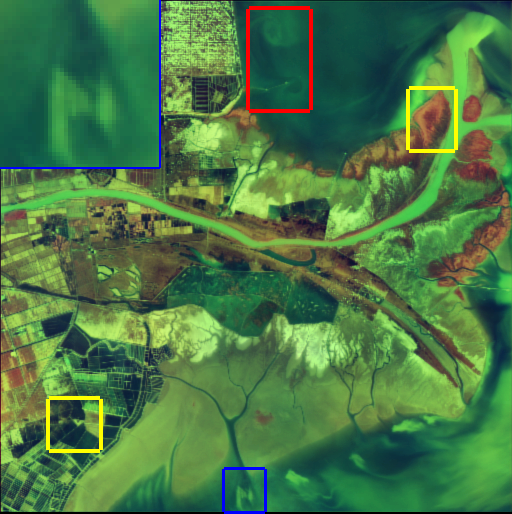}}\!
		
		\vspace{-8pt}
		\subfloat[PMIRFCo]{\includegraphics[width=3.5cm]{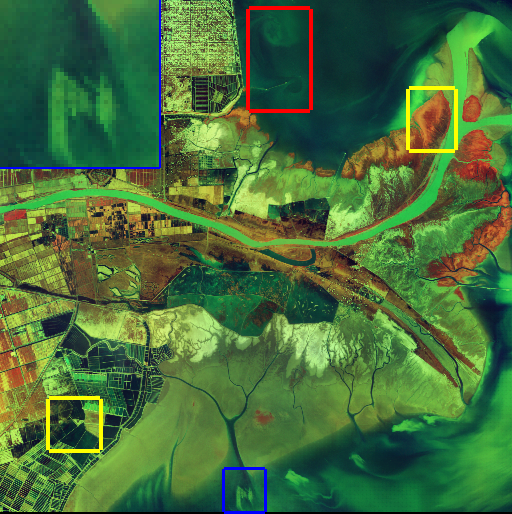}}\!
		\subfloat[DFMF]{\includegraphics[width=3.5cm]{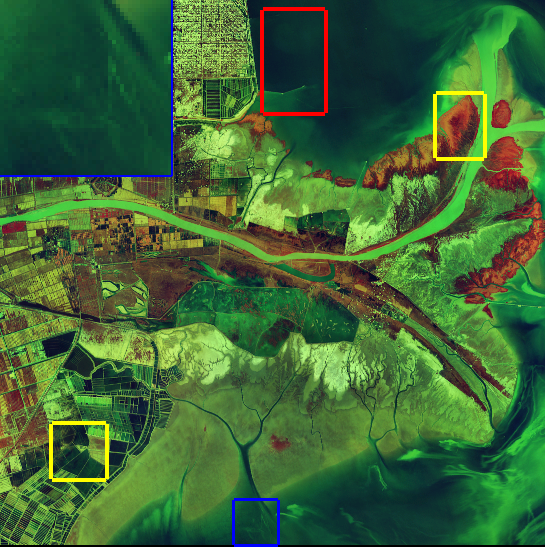}}\!
		\subfloat[HPWRL]{\includegraphics[width=3.5cm]{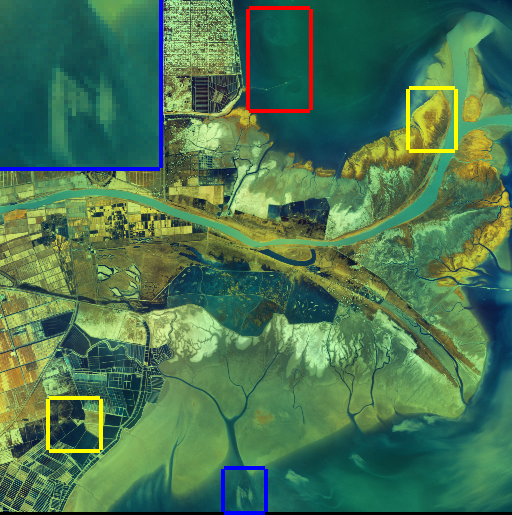}}\!
		\subfloat[IR-ArF]{\includegraphics[width=3.5cm]{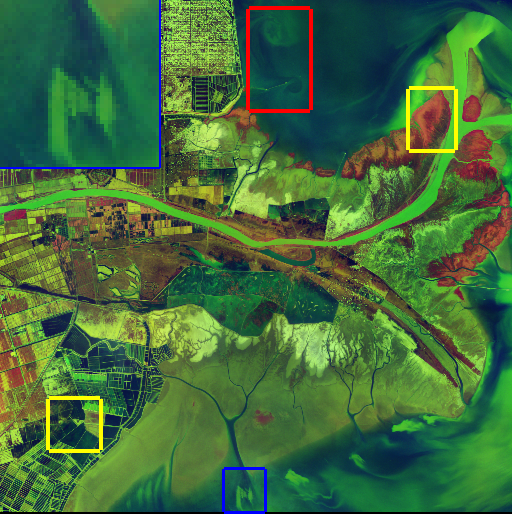}}\!
		\subfloat[SDR-BSF]{\includegraphics[width=3.5cm]{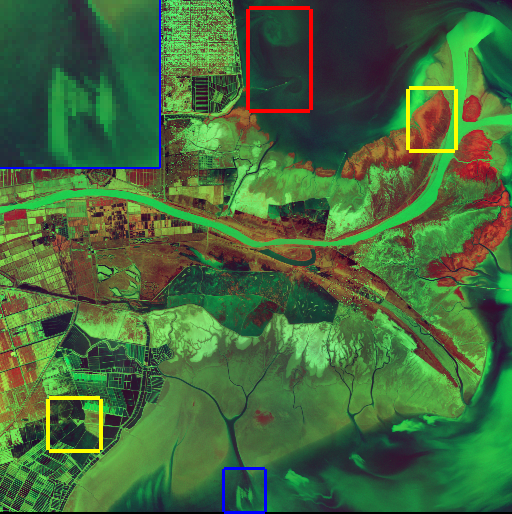}}
		
		\vspace{-2pt}
		
		\caption{Visualization of the fused images  for the GF1-GF5 datasets (bands 30, 60 and 90) using various compared methods. }
		\label{DF0}
		\vspace{-4pt}
	\end{figure*}

	Next, we validate the effectiveness of the proposed SDR-BSF method on real-world datasets: GF1-GF5 and FR2. The image pairs are not pre-registered, and the degradation operators remain unidentified. Due to the absence of a ground truth image, we assess the effectiveness of different methods by  examining image details.  We  display the recovery performance of each compared method on the GF1-GF5 dataset  in Figure \ref{DF0}. 
	
	In the input MSI and the fused images, a `vortex' feature is highlighted with a red square, while a specific pattern detail is indicated by a blue border and magnified in the top-left corner of the image. Meanwhile, yellow squares are used to mark areas for  the comparison of spectral differences.  The `Integrated' method also shows significant color discrepancies, indicating potential biases in the spectral information. Additionally, the fused image obtained from `NED' displays a noticeable sense of discontinuity at the common scene boundaries. In comparison, the reconstruction image obtained by `NonregSR' shows some blurriness and is accompanied by slight color differences.  Intuitively, `PMIRFCo'  achieves superior restoration performance while retaining important image details. However, the enlarged view reveals that the fused image suffers from slight distortion. `HPWRL' performs well in the spatial dimension. However,  there are obvious color differences   compared to the original HSI. The reconstruction images from `DFMF' are obtained by first performing pre-registration of the HSI and MSI using geolocation coordinates, followed by the `DFMF' method. As can be seen, `DFMF' achieves high-quality reconstruction overall. Nevertheless, some textural details in the reconstruction images are smoothed out at the locations marked in red and blue. Visually, `IR-ArF' produces a higher-quality fused image. In contrast, the proposed SDR-BSF method retains important image details while exhibiting less color differences, which validates the effectiveness of our approach on real dataset.

	\begin{figure*}[htbp]
		\captionsetup[subfloat]{labelsep=none,format=plain,labelformat=empty}
		\centering
		\vspace{-10pt}
		\subfloat[HSI]{\includegraphics[width=3.5cm]{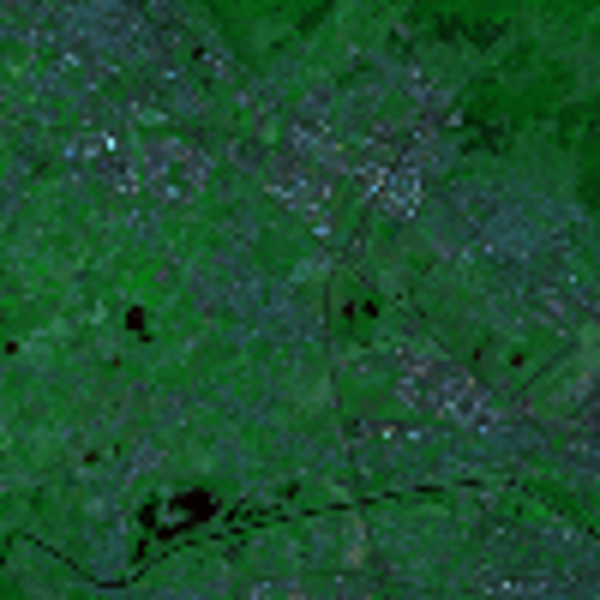}}\!
		\subfloat[Pan]{\includegraphics[width=3.5cm]{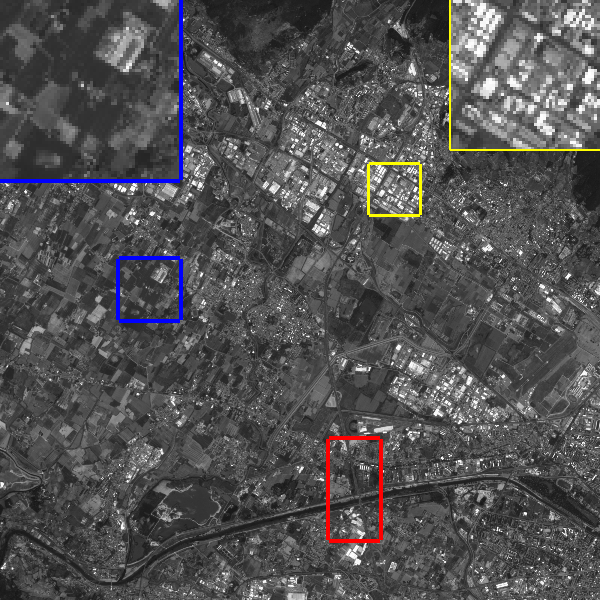}}\!
		\subfloat[Integrated]{\includegraphics[width=3.5cm]{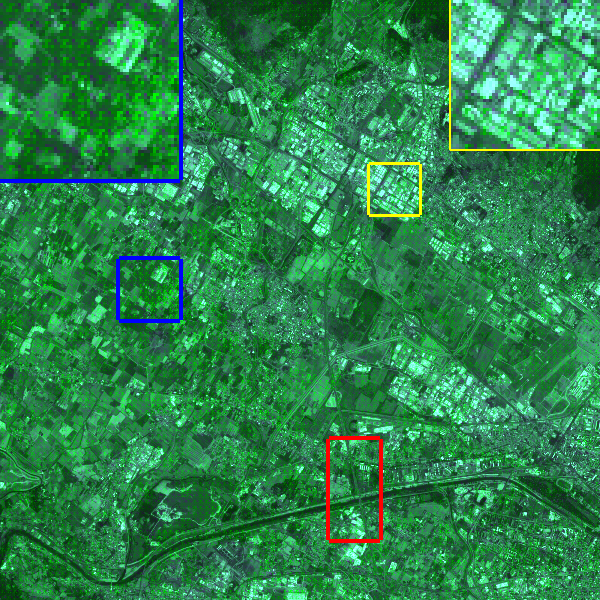}}\!
		\subfloat[NED]{\includegraphics[width=3.5cm]{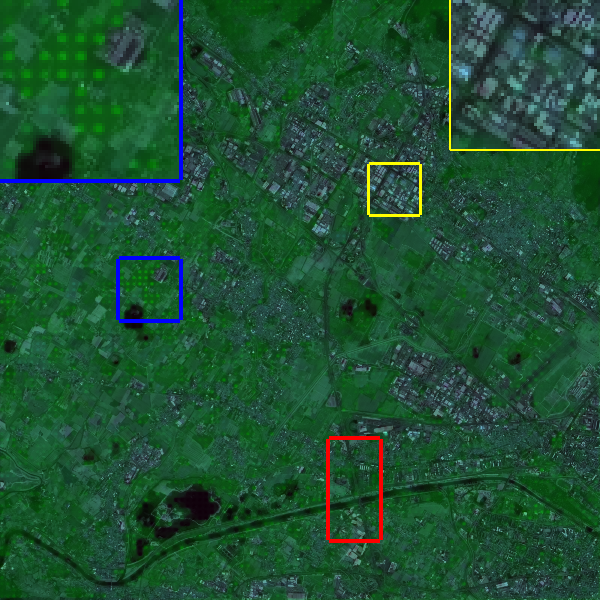}}\!
		\subfloat[NonregSR]{\includegraphics[width=3.5cm]{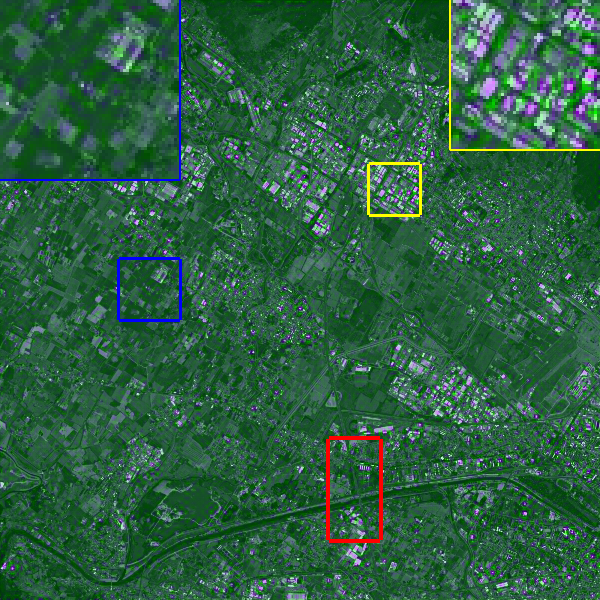}}\!
		
		\vspace{-8pt}
		\subfloat[PMIRFCo]{\includegraphics[width=3.5cm]{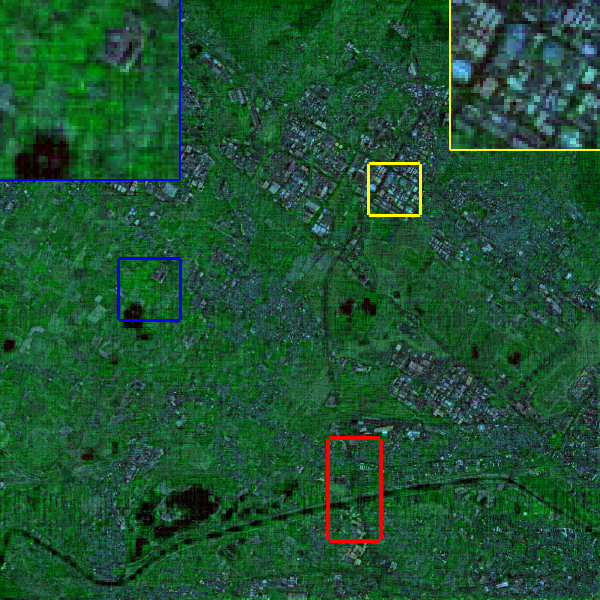}}\!
		\subfloat[DFMF]{\includegraphics[width=3.5cm]{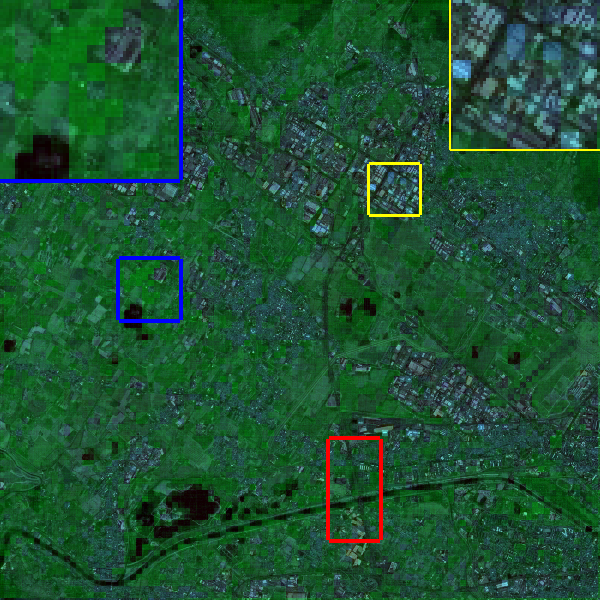}}\!
		\subfloat[HPWRL]{\includegraphics[width=3.5cm]{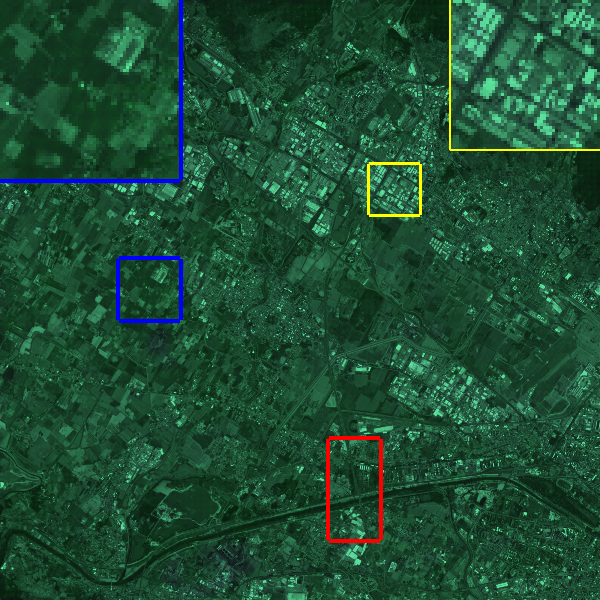}}\!
		\subfloat[IR-ArF]{\includegraphics[width=3.5cm]{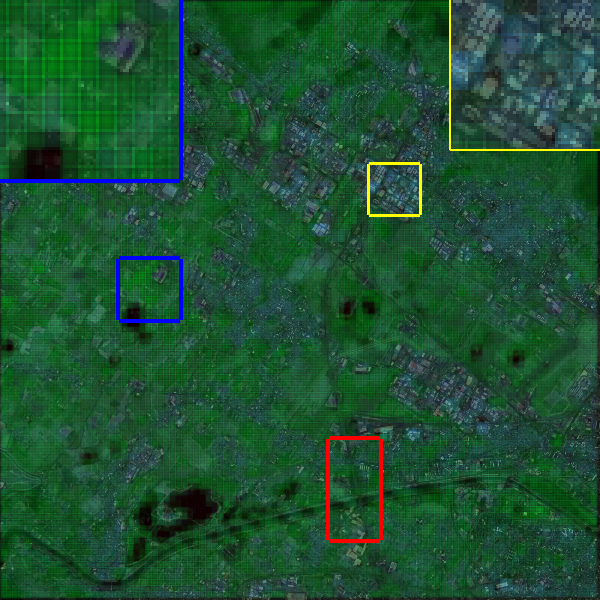}}\!
		\subfloat[SDR-BSF]{\includegraphics[width=3.5cm]{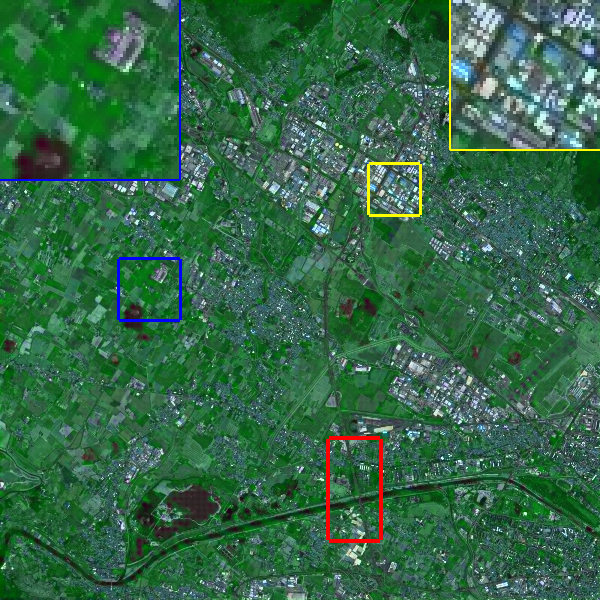}}
		
		\vspace{-0pt}
		
		\caption{Visualization of the fused images  for the FR2 datasets (bands 15, 30 and 45) using various compared methods. }
		\label{DFF0}
		\vspace{-4pt}
	\end{figure*}
	
	Figure \ref{DFF0} displays the restored images from each method on the FR2 dataset, with the blue and yellow regions magnified for closer inspection.  This dataset contains highly complex scenes, which result in relatively noticeable distortions in the fused images across various methods. The methods `NED', `PMIRFCo', and `DFMF' do not exhibit noticeable color differences in their fused images. However, upon closer inspection of the magnified regions, it is evident that these images suffer from varying degrees of spatial distortion. One possible reason for this is the complexity of the scenes in the FR2 dataset, which makes it challenging to achieve high-quality recovery of spatial information. 
	The fused image from `Integrated', `NonregSR' and `HPWRL' exhibit  color differences, and some important details in the red  region have been smoothed out.   The fused image of `IR-ArF' exhibits stripe distortions. In contrast, the fused image obtained from our method, while exhibiting slight color differences, retains important details and displays clearer spatial textures. This further validates the effectiveness of our approach.

	\subsection{Running time}
	Next, we compare the runtime of each algorithm in Table \ref{T5}.
	\begin{table}[hpbt]
		\renewcommand\arraystretch{1.17}
		\caption{The time consumption of the compared methods (unit: seconds).} 
		\vspace{-0.1cm}      
		\centering
		\label{T5}               
		\begin{tabular}{|p{1.8cm}|ccccc|}
			\hline
			\textbf{dataset} & Pavia C  & Superballs & GF1-GF5 & & FR2
			\\ \hline
			Hysure  \cite{Hysure}	& 47  & 69 &  124  & & 84 \\
			Integrated \cite{Zhou2020} & 260  & 373 &  757 && 526 \\ 
			NED  	\cite{NED}	& 24  & 33 &  51 & &44 \\
			NonregSR \cite{Zheng2022} 	& 931 & 1225  & 3560 && 3030\\ 
			PMIRFCo  \cite{PMI}	& 438 & 562 &  1275  &&  913 \\ 
			DFMF	\cite{Guo2022}	& 259  & 501 & 1065 && 845\\ 
			HPWRL  	 \cite{Nie2024}	& 254  & 341 &  863 && 684 \\ 
			IR-ArF  \cite{Qu2025}	& 669  & 854 &  2630 && 2460 \\ 
			SDR-BSF  	&  \textbf{10} & \textbf{13} & \textbf{17} && \textbf{14} \\ 
			\hline
		\end{tabular}
		\vspace{-2pt}
	\end{table}
	From the table,  `Hysure' and `NED' are competitive in terms of computational efficiency since they do not require a training phase. Instead, the majority of their computational time is dedicated to iterative solving processes. The compared  deep learning-based  methods tend to take longer processing time. This is mainly due to the fact that these methods tackle the problem from the spatial dimension, which necessitates the inclusion of complex registration and fusion modules, as well as degradation operator estimation modules. 
	This makes the training process intricate and time-consuming.  In contrast, our method obtains registered images from the perspective of spectral super-resolution, and the designed network is very concise. Moreover, the BSF model does not involve computationally intensive steps. Consequently, the time consumption is significantly lower than that of other comparative methods.

    \subsection{Ablation study}
\begin{figure}[htbp]
		\vspace{-5pt}
		\quad\subfloat{\includegraphics[width=7.7cm,height=5.8cm]{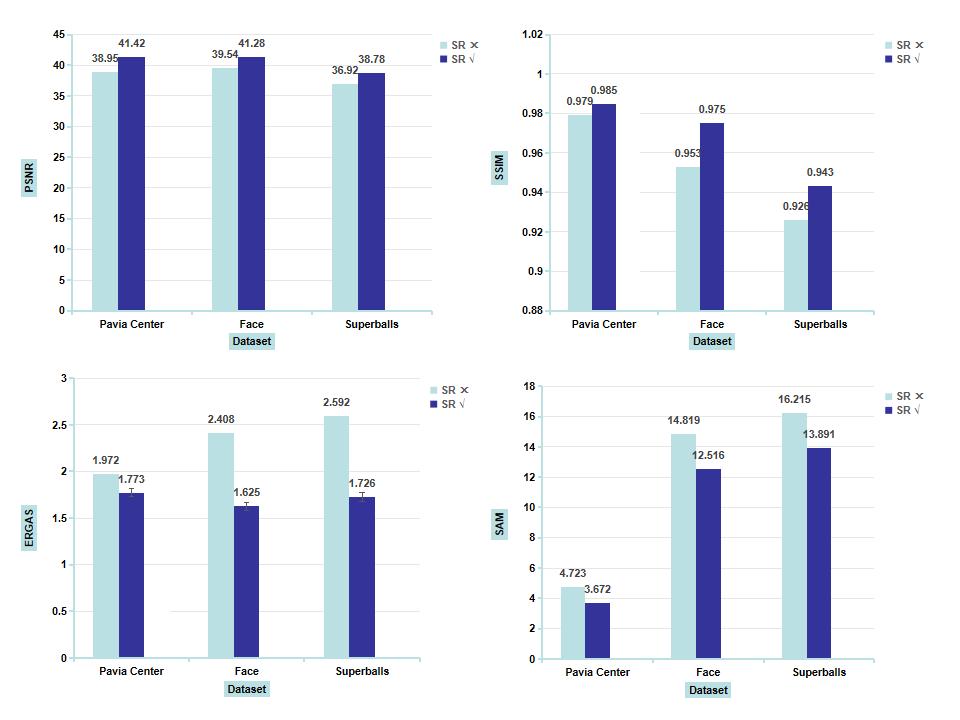}}
		\vspace{-7pt}
		\caption{Ablation experiment for the subspace representation (SR) under \textbf{scaling transformation} for  different HSI datasets.\tiny\label{DF8}}
		\vspace{-1pt}
	\end{figure}

\begin{figure}[htbp]
		\vspace{-5pt}
		\quad\subfloat{\includegraphics[width=7.7cm,height=5.8cm]{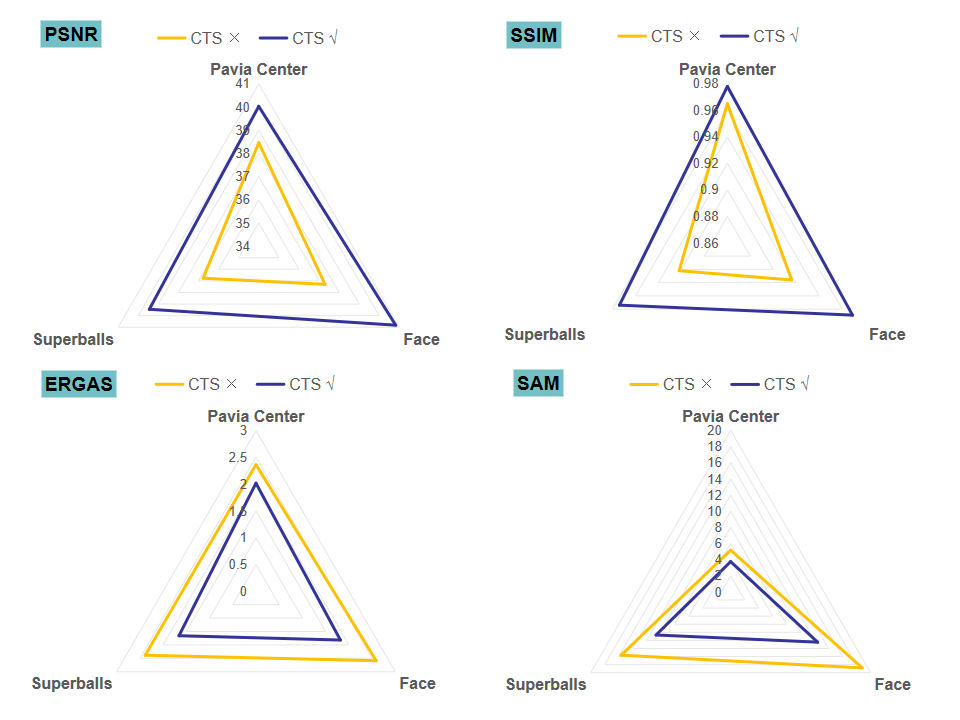}}
		\vspace{-7pt}
		\caption{Ablation experiment for the cyclic training strategy (CTS) under \textbf{pincushion transformation} for different HSI datasets.\tiny\label{DF9}}
		\vspace{-5pt}
	\end{figure}

	Next, we conduct ablation studies to evaluate the effectiveness of our  innovations. For the SPL network, we  validate the efficacy of  the subspace representation (SR) method and the cyclic training strategy (CTS) in Figure \ref{DF8} and \ref{DF9}. 

    Figure \ref{DF8}  demonstrates that the SR approach has  a positive impact on the registration and fusion process process for all the three datasets. For example, it improved the PSNR of the fused image  by nearly 2.5 dB for `Pavia Center' dataset. Other evaluation metrics also showed significant improvements across different datasets, which verifies the effectiveness of the SR method. 
	Figure \ref{DF9} shows that cyclic training strategy  can  decrease the spectral deviation caused by unregistered original image pairs. We focused on the SAM metric and found that CTS showed a  significant improvement for all the three datasets. This indicates that CTS can effectively enhance the spectral reconstruction accuracy of fused images.

    \subsection{The influence of fusion on classification accuracy}
	To demonstrate the effectiveness of the  SDR-BSF approach in subsequent downstream tasks, we evaluate its impact on classification accuracy using the Houston dataset \cite{Houston}. This dataset comprises a spatial resolution of 349 × 1905 pixels and  144 spectral bands. We chose spectral bands 11 through 110 and  average every 25 bands to obtain the MSI. The HSI is generated by applying an 7 × 7 Gaussian kernel with a standard deviation of 2 and a downsampling ratio of 8. Furthermore,  we applied a 2° rotation to the HSI and performed a barrel transformation to misalign it with the MSI. Subsequently, the fused image was obtained using the SDR-BSF model.
	\begin{table}[htbp]\small
		\renewcommand\arraystretch{1.15}
		\begin{center}
			\caption{Classification results of LR-HSI and fused HSI (Accuracy(\%)). }
			\label{class}
			\vspace{-2pt}
			\begin{tabular}{p{2.3cm}|p{1.5cm}p{1.5cm}c}
				\hline
				\textbf{Category} & \textbf{HSI} &\textbf{MSI} & \textbf{Fused HSI} \\
				\hline
				Healthy grass   & 70  &\textbf{92}  & \textbf{92} \\
				Stressed grass  & 67 &\textbf{95} & 94 \\
				Synthetic grass & 71 & \textbf{99} & \textbf{99} \\
				Trees           & 60 &96 & \textbf{97} \\
				Soil 			& 76 & 86 & \textbf{88} \\
				Water 			& 72 &95 & \textbf{95} \\
				Residential 	& 54 & 68 & \textbf{72} \\
				Commercial 		& 72 & 49 & \textbf{74} \\
				Road 			& 66 & 72 & \textbf{72}  \\
				Highway 		& 84 &69 & \textbf{86} \\
				Railway 		& \textbf{84} & 76& 82 \\
				Parking lot 1 	& \textbf{71} &67 & 70 \\
				Parking lot 2 	& 62 &\textbf{63} & \textbf{63} \\
				Tennis court 	& 96 & 93& \textbf{97} \\
				Running track 	& 87 &96 & \textbf{97} \\
				\hline
				Average & 72.8 & 81.1 & \textbf{84.9} \\
				\hline
			\end{tabular}
		\end{center}
		\vspace{-2pt}
	\end{table} 
	Then, we classified the HSI, MSI and the fused image using the SVM-KC algorithm \cite{SVMKC}. Table \ref{class} presents the classification outcomes, highlighting both the accuracy for each category and the overall average accuracy. The classification performance for most categories on the fused HSI surpasses that of the HSI and MSI. Notably, the average accuracy for the predicted HR-HSI is 12.1\% higher than the HSI and 3.8\% higher than MSI. This validates the effectiveness of our  method in enhancing HSI classification performance.
	
	\section{CONCLUSION}\label{G}
	In this paper, we proposed the SDR-BSF method to address the blind fusion challenge of unregistered HSI-MSI pairs. We innovatively propose to perform registration in the spectral domain, achieving HSI registration by applying spectral super-resolution to the MSI.  Through cyclic training and subspace representation methods, we acquire higher quality and more robust registration performance.
	Next,  the semi-blind BSF model is developed to fuse the registered images, which leverages group sparsity to characterize the low-rank property of the image. This approach eliminates the need for computationally expensive SVD. We solve the BSF model using PAO algorithm and  provide its detailed convergence analysis. Finally, extensive numerical experiments are conducted to verify the effectiveness of our approach.

\bibliographystyle{IEEEtran}
\bibliography{SSDR/CiteTex}

\end{document}